\documentclass{article}

\usepackage[final]{neurips_2025}

\usepackage[dvipsnames]{xcolor}

\definecolor{titlecolor}{RGB}{238,130,238}
\newcommand{\ourmethod}{EA3D}

\newif\ifdrafting
\draftingtrue 
\draftingfalse 
\ifdrafting
    \newcommand{\todo}[1]{{\leavevmode\color[rgb]{1,0,0}[TODO: #1]}}

\else
    \newcommand{\todo}[1]{}
    
\fi

\definecolor{tabfirst}{rgb}{1, 0.7, 0.7} 
\definecolor{tabsecond}{rgb}{1, 0.85, 0.7} 
\definecolor{tabthird}{rgb}{1, 1, 0.7} 

\usepackage[utf8]{inputenc} 
\usepackage[T1]{fontenc}    
\usepackage[pagebackref=true,breaklinks,colorlinks,citecolor=blue,linkcolor=blue,urlcolor=blue,hypertexnames=false]{hyperref}
\usepackage{url}            
\usepackage{booktabs}       
\usepackage{amsfonts}       
\usepackage{nicefrac}       
\usepackage{microtype}      
\usepackage{xcolor}         

\usepackage{float}
\usepackage{bbm}
\usepackage{xspace}
\usepackage{wrapfig}
\usepackage{graphicx}
\usepackage{enumitem}
\usepackage{amsmath}
\usepackage{multirow}
\usepackage{pifont}
\usepackage{colortbl}
\usepackage{wrapfig}
\usepackage{subcaption}
\usepackage{amssymb}

\usepackage{amssymb}
\usepackage{bbding}
\usepackage{pifont}
\usepackage{wasysym}
\usepackage{utfsym}
\usepackage{fontawesome}

\definecolor{isabelline}{RGB}{244, 240, 236}
\definecolor{kaiming-green}{RGB}{57,181,74}
\title{\emph{\textbf{\textcolor{titlecolor}{\ourmethod{}}}}: Online Open-World 3D Object Extraction \\ from Streaming Videos}

\author{
~ Xiaoyu Zhou\textsuperscript{1}$^\dagger$
~ Jingqi Wang\textsuperscript{1}$^\dagger$
~ Yuang Jia\textsuperscript{1}
~ \textbf{Yongtao Wang\textsuperscript{1}}\thanks{Corresponding author.} \\
~ \textbf{Deqing Sun\textsuperscript{2}}
~ \textbf{Ming-Hsuan Yang\textsuperscript{2, 3}} \\
{\textsuperscript{1}Wangxuan Institute of Computer Technology, Peking University} \\
~ {\textsuperscript{2}Google DeepMind}
~ {\textsuperscript{3}University of California, Merced} \\
}

\begin{document}

\maketitle


{
\begin{figure}[H]
\hsize=\textwidth
\centering
\includegraphics[width=1.0\linewidth]{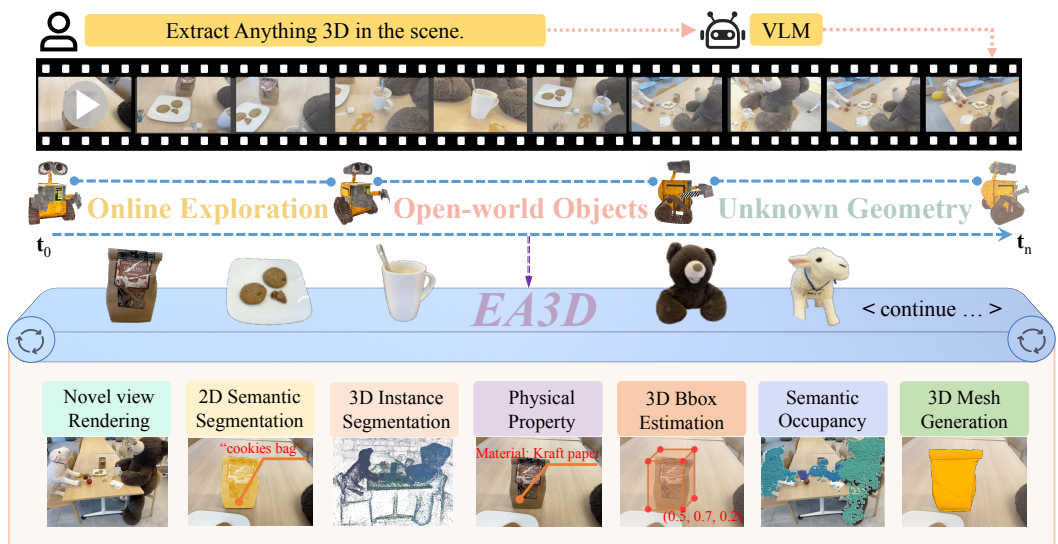}
\caption{Illustration of {\em \textbf{ExtractAnything3D}} ({\textcolor{titlecolor}{\ourmethod{}}}), which enables online open-world 3D object extraction. Given a streaming video as input with unknown geometry, pose, or semantics, \ourmethod{} performs online and simultaneous scene interpretation and geometry reconstruction, enabling multi-task understanding and modeling of any 3D objects in the scene.
}
\label{fig:teaser}
\end{figure}
}

\begin{abstract}
Current 3D scene understanding methods are limited by offline-collected multi-view data or pre-constructed 3D geometry. In this paper, we present {\em \textbf{ExtractAnything3D}} ({\textcolor{titlecolor}{\ourmethod{}}}), a unified online framework for open-world 3D object extraction that enables simultaneous geometric reconstruction and holistic scene understanding. 
Given a streaming video, \ourmethod{} dynamically interprets each frame using vision-language and 2D vision foundation encoders to extract object-level knowledge. This knowledge is integrated and embedded into a Gaussian feature map via a feed-forward online update strategy. We then iteratively estimate visual odometry from historical frames and incrementally update online Gaussian features with new observations. A recurrent joint optimization module directs the model's attention to regions of interest, simultaneously enhancing both geometric reconstruction and semantic understanding.
\footnote{Contribute equally.}
Extensive experiments across diverse benchmarks and tasks, including photo-realistic rendering, semantic and instance segmentation, 3D bounding box and semantic occupancy estimation, and 3D mesh generation, demonstrate the effectiveness of \ourmethod{}. Our method establishes a
unified and efficient framework for joint online 3D reconstruction and holistic scene understanding,
enabling a broad range of downstream tasks. The project webpage is available at \url{https://github.com/VDIGPKU/EA3D}.
\end{abstract}
\section{Introduction}
To see is, as famously defined by David Marr~\cite{marr2010vision}, ``\emph{to know what is where by looking.}''
For an autonomous agent, such as a robot, operating in an unfamiliar environment, this translates into formidable challenges. 
Imagine a robot entering a new room, observing and understanding its surroundings on the fly (Fig.~\ref{fig:teaser}). 
It faces an unknown quantity and variety of objects (\textbf{open world}) and needs to process unfamiliar 3D geometry (\textbf{unknown geometry}) in a streaming mode (\textbf{online exploration}). To effectively navigate and interact within such a dynamic 3D space, the robot must be able to dynamically construct open-world 3D representations of the scene. Concurrently, it must comprehend the geometric structures and physical properties of the objects it encounters and perceptively model the motion states of all semantic entities within complex, evolving environments.

While Vision-Language Models (VLMs)~\cite{hurst2024gpt, zhang2023llama, lin2023video} show impressive results on 2D open-world understanding , they struggle in 3D domains, exhibiting view inconsistencies\cite{zhen20243d, hong20233d}, geometric misalignment\cite{asano2025geometry}, and inability to handle occlusions.
A straightforward solution is to lift 2D VLM outputs into 3D using scene geometry~\cite{yang2023sam3d, zhang2023sam3d, jiang2024open}, but this requires pre-constructed 3D geometry, annotated datasets for training, and still suffers from 3D-2D misalignment issues. Recent differentiable rendering frameworks like NeRF~\cite{mildenhall2021nerf, ren2024nerf} and 3DGS~\cite{kerbl20233d, yu2024mip, fu2024colmap} enable joint 3D scene understanding by optimizing 3D representations with pixel-level pseudo-labels\cite{kerr2023lerf, ye2024gaussian, zuo2025fmgs, qin2024langsplat}. However, these offline approaches require complete multi-view images and time-consuming multi-stage processes.

In this paper, we introduce {\em \textbf{ExtractAnything3D}} ({\textcolor{titlecolor}{\ourmethod{}}}), an online open-world scene understanding framework that simultaneously explores, reconstructs, and interprets the 3D geometry and semantic knowledge of a scene. Similarly to human perception, our system starts processing streaming visual inputs as soon as it enters a room, reconstructing and understanding the current scene online based on historical observations and prior knowledge. As new frames emerge, they progressively reveal more comprehensive spatial information, enriching the internal knowledge base and allowing the system to infer occluded regions via novel view synthesis.
Specifically, we utilize VLMs to openly interpret object categories and physical properties from the emerging frame while dynamically maintaining a semantic cache. We then combine features from multiple visual foundation models with semantic cues to construct a dynamically updated knowledge-integrated feature map. The knowledge-integrated features are embedded into Gaussian representations through a fast feedforward step and are updated jointly over time. To incrementally extract both geometry and knowledge of 3D objects in an online manner, we construct Online Feature Gaussians, consisting of two core components: online visual odometry and online Gaussian updating. Benefiting from a recurrent joint optimization strategy, our proposed Online Feature Gaussians dynamically extract any 3D objects in the scene, facilitating multiple tasks including photo-realistic rendering, semantic and instance segmentation, physical property analysis, and geometric reasoning (e.g., 3D bounding boxes, semantic occupancy, and 3D mesh generation). \ourmethod{} thus establishes a unified and efficient framework for joint online 3D reconstruction and holistic scene understanding, enabling a wide range of downstream tasks.

The contributions of this work are: 1) We propose a unified online open-world 3D objects extraction framework enabling simultaneous online reconstruction and understanding without geometric or pose priors. 
2) Taking streaming video as input, our method effectively leverages historical knowledge to guide 3D object extraction at the current observation, enabling online joint updates of integrated features and delivering high-quality, efficient geometric reconstruction and scene understanding.
3) Our method supports a broad set of tasks, including photo-realistic reconstruction and rendering, semantic and instance segmentation, 3D bounding box construction, semantic occupancy estimation, and 3D mesh generation, consistently achieving good performance across multiple benchmarks.

\section{Related Work}

\textbf{Open-World Foundation Model.}
When exploring the real world, the quantity and categories of 3D objects remain unknown in unbounded environments. Recent advances in Vision-Language Models (VLMs) and Vision Foundation Models (VFMs) have significantly advanced open-world interpretation of 2D images. VLMs~\cite{hurst2024gpt, lin2023video, zhang2023llama, wang2024cogvlm} effectively fuse visual and textual cues for Visual Question Answering (VQA), while SAM-based~\cite{kirillov2023segment, ravi2024sam} and CLIP-based methods~\cite{gao2024clip, lin2024generative, yao2024detclipv3} excel in generalized semantic segmentation and instance detection.
However, these methods suffer from severe multi-view inconsistencies and semantic ambiguities, especially for small objects, due to their limited geometric awareness. They also struggle with spatial occlusions and suffer from memory degradation over time.
To overcome these challenges, we propose an online, synchronized framework for joint reconstruction and understanding, where 2D foundational features are implicitly aligned throughout the online reconstruction process. Our framework leverages online embedding from VFMs and recurrent joint optimization to seamlessly align 2D knowledge with 3D geometry, ensuring coherent consistency across the 3D domain.

\textbf{3D Scene Understanding.}
Current 3D scene understanding methods broadly categorized into two groups: (1) methods that operate on known 3D geometry—such as point clouds, depth maps, or meshes; and (2) methods that infer scene semantics while reconstructing the 3D geometry.
Methods like~\cite{peng2023openscene, takmaz2023openmask3d} and~\cite{zhai2025panogs, chacko2025lifting} extract semantics via 2D-to-3D lifting, but all depend on pre-built 3D geometry and costly semantic annotations.
Recent approaches address this limitation by jointly reconstructing and segmenting 3D scenes through differentiable rendering. NeRF~\cite{kerr2023lerf, cen2023segment} and 3DGS-based methods~\cite{ye2024gaussian, zuo2025fmgs, qin2024langsplat, li2024instancegaussian} leverage pseudo-labels to jointly optimize appearance and semantics via 2D supervision.
However, both types of methods are inherently offline, relying on full scene observations before reconstruction and interpretation.
In real-world settings, agents dynamically explore and progressively understand scenes. To address this gap, we propose an online framework for simultaneous scene reconstruction and understanding. Our method efficiently builds 3D objects while delivering high-quality semantic interpretation. Guided by evolving 3D geometry, it enables comprehensive extraction of open-world objects.

\textbf{Online Reconstruction.}
Recent advances in 3DGS~\cite{kerbl20233d, yu2024mip} have demonstrated remarkable capabilities in photo-realistic rendering and have been extended to a range of downstream applications, including robotic manipulation~\cite{zheng2024gaussiangrasper, lu2024manigaussian, shorinwa2024splat}, dynamic scene reconstruction~\cite{wu20244d, zhou2024drivinggaussian, huang2024sc, yang2023real}, and 3D content generation~\cite{chung2023luciddreamer, zhou2024gala3d, ren2023dreamgaussian4d}. However, vanilla 3DGS requires prolonged optimization and offline training with access to full video sequences, limiting its practicality in real-world scenarios.

To address these limitations, recent methods~\cite{sun20243dgstream, gao2024hicom, yan2025instant, li2025streamgs} have proposed streaming extensions of 3DGS that significantly reduce training time and memory consumption.
However, they rely on multi-view videos and pre-computed global poses, which are often impractical in real-world settings. SLAM-based approaches~\cite{matsuki2024gaussian, li2025monogs++} also enable online scene reconstruction but rely on sparse keyframe tracking and expensive post-refinement, limiting their ability to capture fine-grained geometry and semantics. In a related effort, an online Gaussian-based method~\cite{wu2024embodiedocc} has been proposed for scene occupancy prediction. However, it is tailored for a specific task, fails to achieve photo-realistic rendering, and suffers from prohibitively expensive training costs. 
To overcome these challenges, we propose a novel online Gaussian optimization strategy based on knowledge feature guidance, enabling joint reconstruction and understanding of scenes in an on-the-fly manner.

\section{Method}

\begin{figure}[t] 
\centering
\includegraphics[width=1.0\textwidth]{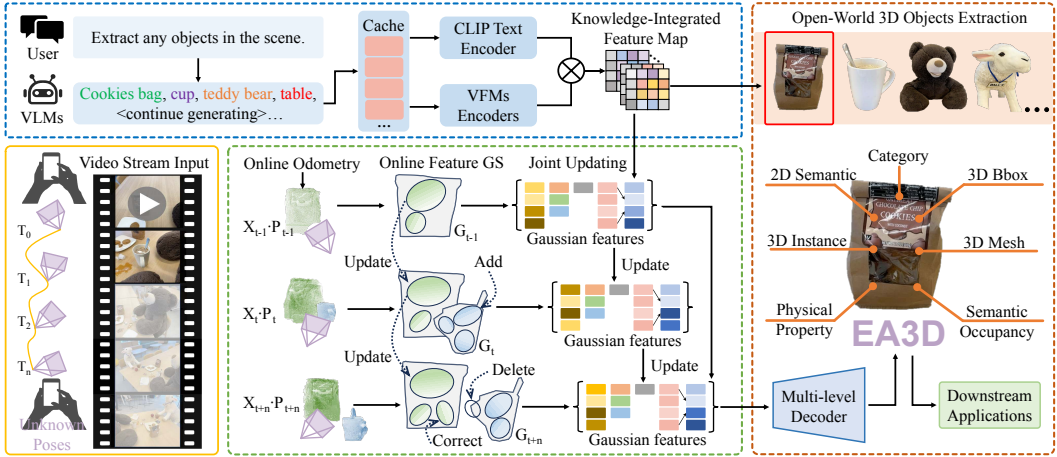}
\caption{\textbf{Framework of \ourmethod{}.} 
Given a streaming video without poses or labels, \ourmethod{} first leverages VLMs to identify all potential objects and their physical attributes, while maintaining a dynamic semantic cache to track newly emerging categories. We then use multi-level VFMs to extract knowledge-integrated feature maps from each frame and embed them into Gaussian primitives via a feedforward way. We perform online visual odometry estimation, and incrementally reconstruct geometry and infer knowledge through our online feature Gaussians. A recurrent joint optimization fuses current observations with historical features to continuously update the Gaussians. \ourmethod{} supports a wide range of 3D perception tasks and shows strong potential for downstream applications.
}
\label{fig:pipeline}
\vspace{-10pt}
\end{figure}

As shown in Fig.~\ref{fig:pipeline}, the proposed ExtractAnything3D (\ourmethod{}) enables open-world 3D object extraction through three key components: \textbf{(a)} Knowledge extraction and integration, leveraging VLMs and multi-level VFMs for open-world understanding, integrating knowledge feature maps with an online cache and dynamically embed them into Gaussians via a feedforward way (Sec~\ref{framework:feature}). \textbf{(b)} Online visual odometry for fast pose estimation and geometric initialization, along with online feature Gaussians that incrementally reconstruct object geometry and transfer knowledge online (Sec~\ref{framework:online}). \textbf{(c)} Joint optimization that continuously updates 3D object representations by fusing current observations with historical features (Sec~\ref{framework:optimization}). \ourmethod{} supports a wide range of 3D tasks.

\subsection{Knowledge-Integrated Feature Map}
\label{framework:feature}

Given a streaming video, we first extract object-level knowledge by dynamically interpreting the scene frame by frame using 2D vision foundation models (VFMs). However, current 2D foundational vision models lack geometric awareness of 3D scenes, leading to significant multi-view inconsistencies and ambiguities, especially in occluded regions. To tackle this challenge, we propose implicitly aligning foundational visual features in 3D space through a multi-view reconstruction pipeline based on Gaussian Splatting (GS). Each 3D representation primitive is embedded within a knowledge-integrated feature map, utilizing a feed-forward online update strategy.

\textbf{Open-world interpretation by VLMs.} VLMs~\cite{hurst2024gpt, zhang2023llama, wang2024cogvlm} have shown exceptional open-world understanding in 2D images. Given an image $I$ observed at timestep $t$, we first use VLMs to identify all instances and their semantics within the image. In an open-world scene, the number and categories of objects are unknown. We use the prompt ``Find and list all the possible objects in the given image'' to capture any potential objects. Considering the continuously evolving number and semantics of objects in a streaming video, we dynamically maintain an online semantic cache $\Omega$.
The online semantic cache takes input of class prompts from VLMs of the current frame, updates the semantics of newly emerged objects, and embeds them into a continuous vector $ T \in \mathbb{R}^{1 \times V} $ using a pretrained text encoder from CLIP~\cite{Zhou_2022_CVPR, yu2023turning}, where $V$ denotes the changeable dimension of the vector space.

\textbf{Semantic feature map.} Despite VLMs providing comprehensive open-world interpretation, they exhibit poor visual localization ability. To address this, we leverage foundational vision models~\cite{gao2024clip, oquab2023dinov2, ravi2024sam} to obtain pixel-level segmentation masks and visual features. Given a newly observed image and the online semantic cache, we utilize a pretrained CLIP visual encoder~\cite{gao2024clip} and the Grounded-SAM encoder~\cite{ren2024grounded} to generate pixel-wise latent visual feature representations corresponding to each semantic. However, these features contain non-negligible noise and redundant information, which interfere with instance-level segmentation. Therefore, we compute the similarity of each category with semantic features using the embedded continuous vector, generating a binary mask for each category.
This mask is then used to aggregate the extracted features using k-nearest neighbors. We then normalize and integrate the semantic features $\mathbf{S} = T \times \mathbf{f}_{sem}$ from different encoders, $\mathbf{f}_{sem}$ denotes the embedded semantic features, and update them into the online semantic cache.

\textbf{Physical Property.} Based on the online semantic cache and 2D priors from VLMs, we also enable the analysis of objects' physical properties. Inspired by~\cite{shuai2025pugs, feng2024pie}, we extend the text prompts to extract object-level and part-level physical properties from VLMs, corresponding to the previously obtained semantics. We then encode the physical attribute features as a variable-length vector $\mathbf{Y}$ with a learnable prompt $y_1, \ldots,y_n$, and fuse it into the online semantic cache. 

\textbf{Feature map embedding.} Vanilla Gaussian Splatting~\cite{kerbl20233d} represents the geometry through a collection of GS parameters, including position $\mu$, covariance matrix $\Sigma$, opacity $o$, and spherical harmonics coefficients to represent appearance. To synchronize the constructing and understanding of the 3D objects, we add an additional knowledge-integrated feature to each Gaussian. Our method integrates VLM priors, foundational visual features, and inter-track cues, combining the strengths of both appearance and geometry. Specifically, we employ a fast feedforward step to embed the knowledge features encoded by visual foundational models into the Gaussian representations. Retrieved from the online semantic cache and dynamically updated, these knowledge features exchange information across streaming frames over time. Given an emerging video frame $I_t$ at time $t$, the integrated knowledge feature map $\mathbf{F}_t^{map}$ can be formulated as:
\begin{align}
\mathbf{F}_t= \sum_{i \in N,j \in N} \mathbf{X^{self}_{i,j}} \cdot \mathbf{S_{i, j}}(\mathbf{T_k} \,; \mathbf{Y_{i, j}} ) \cdot \mathbf{C_t},
\end{align}
where $\mathbf{F}_t$ is the integrated feature map of current frame $I_t$, $\mathbf{S_{i, j}}$ denotes the semantic features and $\mathbf{T_k} \,; \mathbf{Y_{n}}$ are semantic category and physical property tags.
$i,j$ denote the pixel coordinates, $\mathbf{X^{self}_{i,j}}$ and $\mathbf{C_t}$ represent the corresponding point map and confidence map, as introduced in~\ref{framework:online}. Inspired by~\cite{xu2022gmflow}, we then compute the matching distributions of two consecutive video frames:
\begin{align}
\mathbf{M}_{t,t-1} = \text{Softmax}(\frac{{\mathbf{F_t} \mathbf{F_{t-1}}^{\mathsf{T}}}}{\|\mathbf{F_t}\|\|\mathbf{F_{t-1}}^{\mathsf{T}}\|}) ,
\end{align}
where $\mathbf{F_t}, \mathbf{F_{t-1}} \in \mathbb{R}^{H \times W \times D} $ are the feature maps of two adjacent keyframes, where $H$, $W$ and $D$ denote height, width and feature dimension, respectively. $\mathbf{M}_{t,t-1} \in \mathbb{R}^{H \times W \times H \times W}$ is the matching distribution between two adjacent keyframes. Based on the guidance from the matching distributions, we continuously propagate the Gaussian features from the previous view to the current frame via a single forward warping, along with their corresponding knowledge feature maps. This ensures the continuity of knowledge transfer through a simple yet effective forward Gaussian transformation.
We further provide a detailed comparison of our knowledge-integrated feature embedding against existing feature Gaussian methods~\cite{qiu2024feature, zhou2024feature, zhou2025feature4x} in the Appendix.

\textbf{Multi-level decoder for downstream tasks.} Benefiting from the knowledge-integrated feature map, the Gaussian features achieve a unified representation of object geometry and semantics. We then employ a multi-level decoder to decode the Gaussian primitives into diverse outputs, including appearance (i.e., RGB), semantics, physical properties, 3D position, depth map, 3D bounding boxes, and semantic occupancy.

\subsection{Online 3D Objects Extraction}
\label{framework:online}

Suppose we are walking into a room—the construction and understanding of the 3D space begin the moment we step inside and continuously evolve as we explore. To enable this capability, we propose online feature Gaussians, which support incremental extraction of both geometry and knowledge of 3D objects in an online manner.
This framework comprises two core components: 1) \textbf{Online visual odometry}, which iteratively generates and updates the poses as new frames are observed; 2) \textbf{Online Gaussian updating}, which leverages past observations to rapidly reconstruct and understand the current scene, while dynamically correcting previous misconceptions based on new observation. 

\textbf{Online Visual Odometry.} Given an RGB video stream $\{I_t\}_{t=0}^{N}$ without camera pose, we first incrementally estimate the camera pose of the current frame based on a regression of the keypoint graph $(\mathcal{V}, \mathcal{E})$. Each graph node $\mathcal{V}_t$ corresponds to the frame $I_t$ at timestep $t$, and contains the 6-DoF pose $P_t$, pointmap $X_t$, and inverse depth $D_t$. The graph edges $\mathcal{E}$ denotes the correlation between the current frame and historical frames, with corresponding confidence maps $C_t$. We use Cut3R~\cite{wang2025continuous}, a learning-based odometry method, in combination with~\cite{lipson2024deep} to estimate the initial pointmap and confidence map. Unlike concurrent work~\cite{matsuki2024gaussian, li2025monogs++}, we integrate the dense pixel-level point map generated by Cut3R with sparse points from~\cite{lipson2024deep} to more effectively capture the tiny objects in the scene. However, the poses estimated by Cut3R introduce noticeable biases and errors, which accumulate over time. Therefore, we maintain an online keypoint graph and iteratively update it during reconstruction as new frames are processed. Inspired by the local bundle adjustment optimization~\cite{mouragnon2009generic} problem, we use a cost function adopted from~\cite{chen2023dbarf} over the keypoint graph to minimize the reprojection error and update poses for the current frame.

\textbf{Online Gaussian Updating.}
Streaming video enables dynamic observation of 3D objects through continuously emerging views, allowing previously under-observed regions to be completed and occlusion-induced ambiguities to be resolved. Inspired by this, we incrementally add feature Gaussians per frame to refine existing geometry and extract new objects. Our approach builds upon HiCoM~\cite{gao2024hicom}, a streaming GS method designed for multi-view video reconstruction, but overcomes its reliance on predefined poses and multi-view inputs, making it suitable for fully online settings while addressing geometric and semantic challenges.

\begin{figure}[t] 
\centering
\includegraphics[width=0.95\textwidth]{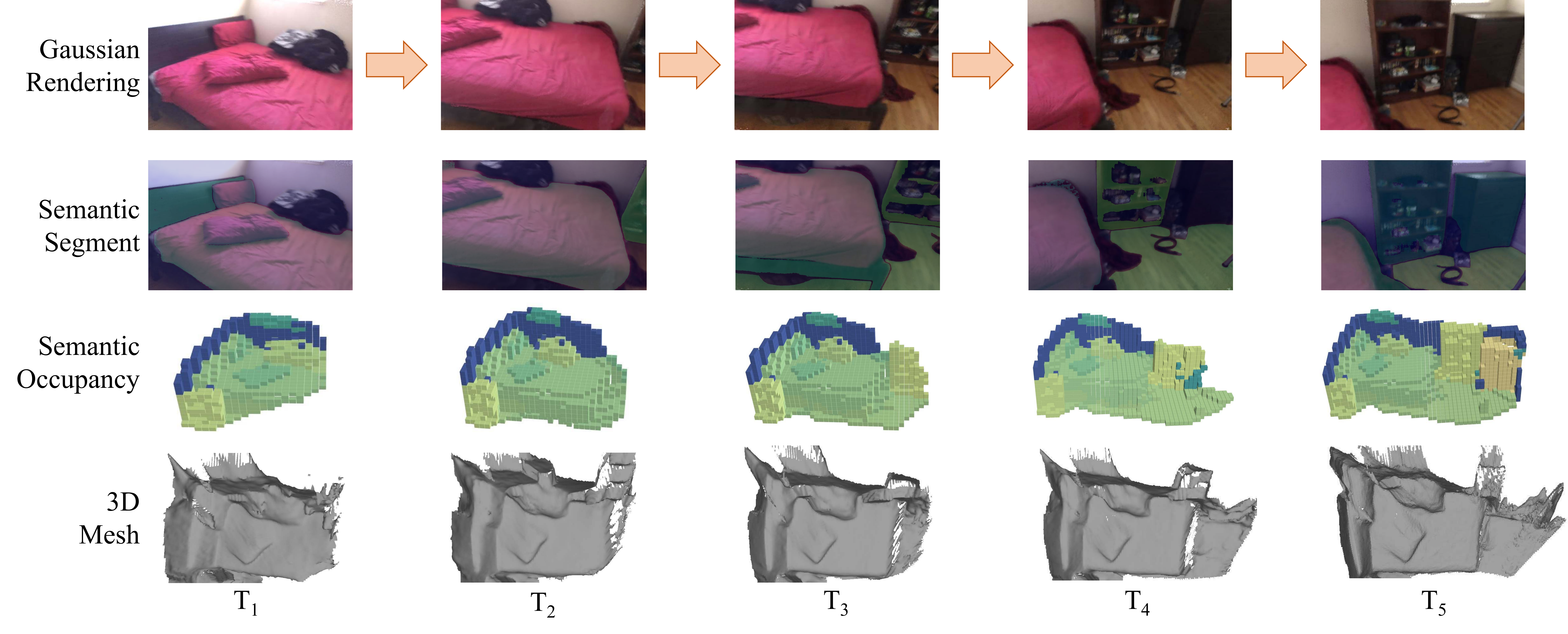}
\caption{\textbf{Visualization of online Gaussian on Scannet~\cite{dai2017scannet}.} \ourmethod{} processes streaming video to incrementally reconstruct while understanding. Historical features guide fast reasoning of current semantics and geometry, while new observations recurrently refine ambiguities and occlusions.
}
\label{fig:online}
\vspace{-10pt}
\end{figure}

To overcome these limitations, we develop a semantics-aware online Gaussian update strategy that incrementally adds and adjusts Gaussians based on historical memory and current observations.
We initialize Gaussians at timesteps 0 and 1. For each new frame, we back-project the online-estimated inverse depth map $D_t$ and pointmap $X_t$ into 3D to obtain an initial point cloud $\Phi$ for object $O \in \Omega$, which is used to initialize the corresponding Gaussians. To reduce redundancy, we adopt the transition strategy from~\cite{gao2024hicom, sun20243dgstream}, assigning each Gaussian a shared translation vector and rotation quaternion within co-visible regions to maintain inter-frame consistency. For newly observed areas, we introduce new Gaussians with means $\mu_i$ initialized from the point cloud, while other attributes are optimized directly. Due to changes in occlusion, some high-opacity ellipsoids may emerge that no longer contribute to specific 3D objects, and we remove them accordingly.
Additionally, we apply a one-step splitting strategy to enable adaptive Gaussian growth based on gradients, improving the representation of under-reconstructed regions. Gradients from the entire scene are finally backpropagated to jointly optimize both Gaussian parameters, features and camera poses.

\subsection{Recurrent Joint Optimization} 
\label{framework:optimization}
During online 3D object extraction, geometric reconstruction and scene understanding mutually reinforce each other. Scene knowledge priors guide the model to focus on areas of interest, while detailed geometry aids in correcting spatial inconsistencies in the priors. Notably, our method enables online joint optimization, without the need for additional post-refinement~\cite{matsuki2024gaussian, li2025monogs++}.

\textbf{Semantic-aware adaptive Gaussian.} To leverage the correlation between object semantics and geometry, we design an adaptive semantic-awareness regularization to guide Gaussian scale adjustment:
\begin{align}
\mathcal{L}_{\mathbf{\delta}} = \sum|\delta_i - \bar{\delta}|F_{sem}^q,
\end{align}
where $\delta_i$ is the scale of the $i$-th Gaussian, and $\bar{\delta}$ is the mean scale of the particular semantic Gaussians,
$F_{sem}^q$ denotes the semantic feature map corresponding to the $q$-th object in the semantic cache $\Omega$.
The semantic-awareness regularization term encourages Gaussians of the same category to share similar scales,
thereby reducing computational overhead caused by redundant scales. After optimizing the integrated Gaussian features, we perform alpha-blending to accumulate the final splatted feature $\hat{F}$: 
\begin{align}
\hat{F} = \sum_{i \in N} F_i \cdot \alpha_i \prod_{j=1}^{i-1}(1-\alpha_j) , 
\end{align}
where $\alpha_i$ denotes the opacity, $F_i$ is the integrated feature map of the $i$-th Gaussian.

\textbf{Joint Semantic-geometry Optimization.} During online Gaussian training, we jointly optimize Gaussian features and camera poses using a combination of photometric loss, geometric loss, knowledge-integrated loss, and regularization terms, formulated as:
\begin{align}
\mathcal{L} = \sum_{t=0}^{t_{now}} \lambda_1\mathcal{L}_1 + \lambda_2 \mathcal{L}_d + \lambda_3\mathcal{L}_{kw} + \mathcal{L}_{\mathbf{\delta}},
\end{align}
where $\mathcal{L}_1$ is the $L_1$ photometric loss. $\mathcal{L}_d = \sum |\hat{D_t} - D_t|$, where $\hat{D_t}$ denotes the rendered depth from Gaussian splatting. $\mathcal{L}_{kw}$ denots the $L_2$ distance between knowledge-integrated feature map and rendered feature map. $\lambda_1$, $\lambda_2$, and $\lambda_3$ are the weighting factors to balance the loss terms. $t_{now}$ denotes the current time step and $t_{0}$ is the initial frame. The loss is dynamically computed on the current frame to update existing Gaussian parameters and features, while future frames remain unseen.

\section{Experiments}
\label{sec:exp}

\setlength\tabcolsep{1.5pt}
\begin{table*}
\small
\caption{Comparison results on ScanNet~\cite{dai2017scannet}. The best results are highlighted in \textbf{bold}, and the second-best results are \underline{underscored}.
``$*$'' indicates the use of the colmap-estimated poses following~\cite{ye2024gaussian, qin2024langsplat, qiu2024feature}. ``$-$'' indicates that the method does not support the specified task. ``Rec., Seg., Bbbox., Occ.'' denotes four multi-task evaluations: reconstruction quality, instance segmentation, 3D bounding box estimation, and semantic occupancy estimation.
}
\begin{tabular*}{\linewidth}{@{\extracolsep{\fill}} lccc | cc | cc | cc | cc}
\toprule
\multicolumn{4}{c}{Tasks:} & \multicolumn{2}{c}{Rec.} & \multicolumn{2}{c}{Seg.} & \multicolumn{2}{c}{Bbox.} & \multicolumn{2}{c}{Occ.} \\
\midrule
Method & Input & Online & Pose-free & PSNR & SSIM & mIoU & mAcc & AP & mAP & IoU & mIoU \\ 
\midrule
LangSplat~\cite{qin2024langsplat} & RGB & \textcolor{red}{\usym{2717}} & \textcolor{red}{\usym{2717}} & 18.4  & 0.69  & 27.5  & 51.3 & -  & -  & -  & - \\
GaussianGrouping~\cite{ye2024gaussian} & RGB & \textcolor{red}{\usym{2717}} & \textcolor{red}{\usym{2717}} & 19.6  & 0.74  & 32.6  & 56.9  & 43.6  & 24.5  & 47.4  & 22.1 \\
FeatureGS~\cite{qiu2024feature} & RGB & \textcolor{red}{\usym{2717}} & \textcolor{red}{\usym{2717}} & 23.9  & 0.84  & 41.1  & 66.0  & 51.4  & 32.7  & 50.9  & 31.2 \\
OpenGaussian~\cite{wu2024opengaussian} & RGB & \textcolor{red}{\usym{2717}} & \textcolor{red}{\usym{2717}} & 22.1  & 0.80  & 35.4  & 61.7  & 47.5  & 28.2  & 49.1  & 25.3 \\
InstanceGaussian~\cite{li2024instancegaussian} & Points & \textcolor{red}{\usym{2717}} & \textcolor{red}{\usym{2717}} & 24.5  & 0.83  & 40.5  & 65.7  & 52.3  & 33.4  & 53.5  & 32.8 \\
\midrule
OpenScene~\cite{peng2023openscene} & Points & \textcolor{ForestGreen}{\usym{2713}} & \textcolor{red}{\usym{2717}} & -  & -  & 42.8  & 68.6  & 55.7  & 34.8  & 51.8  & 30.5 \\ 
EmbodiedSAM~\cite{xu2024embodiedsam} & RGB-D & \textcolor{ForestGreen}{\usym{2713}} & \textcolor{red}{\usym{2717}} & -  & -  & 44.2  & 71.4  & \underline{58.1}  & 39.5  & \underline{55.2}  & 33.0 \\
SAM3D~\cite{yang2023sam3d} & Points & \textcolor{ForestGreen}{\usym{2713}} & \textcolor{red}{\usym{2717}} & -  & -  & 39.2  & 62.3  & 53.7  & 29.1  & 53.3  & 26.7 \\
\midrule
Enhanced Baselines: \\
\midrule
HiCOM~\cite{gao2024hicom}+VFM~\cite{ren2024grounded} & RGB & \textcolor{ForestGreen}{\usym{2713}} & \textcolor{red}{\usym{2717}} & 22.6  & 0.82  & 34.8  & 61.9 & 52.5  & 23.8  &  42.4  & 27.9 \\
MonoGS~\cite{matsuki2024gaussian}+VFM~\cite{ren2024grounded} & RGB & \textcolor{ForestGreen}{\usym{2713}} & \textcolor{red}{\usym{2717}} & 24.3  & 0.85  & 36.3  & 60.5 & 51.7 & 27.7 &  44.5  & 27.2 \\
EmbodiedOcc~\cite{wu2024embodiedocc}+$\mathcal{L}_{RGB}$ & RGB & \textcolor{ForestGreen}{\usym{2713}} & \textcolor{red}{\usym{2717}} & 17.6  & 0.65 & 29.2  & 54.8  & 56.2  & 35.6  & 54.6  & 33.1 \\
FeatureGS~\cite{qiu2024feature}+HiCOM~\cite{matsuki2024gaussian} & RGB & \textcolor{ForestGreen}{\usym{2713}} & \textcolor{red}{\usym{2717}} & 24.5  & 0.85  & 40.8  & 66.3 & 55.8  & 34.7  &  50.7  & 31.4 \\
\midrule
\textbf{\ourmethod{}$^*$} & RGB & \textcolor{ForestGreen}{\usym{2713}} & \textcolor{red}{\usym{2717}} & \underline{25.5}  & \underline{0.87}  & \underline{45.9}  & \underline{71.2}  & \textbf{59.2}  & \underline{39.6}  & 55.0  & \textbf{34.3} \\
\textbf{\ourmethod{}} & RGB & \textcolor{ForestGreen}{\usym{2713}} & \textcolor{ForestGreen}{\usym{2713}} & \textbf{25.8}  & \textbf{0.89}  & \textbf{46.3}  & \textbf{71.8}  & 57.9  & \textbf{39.9}  & \textbf{55.4}  & \underline{33.9} \\
\bottomrule
\end{tabular*}
\label{tab:comparison}
\vspace{-10pt}
\end{table*}

\textbf{Datasets.}
We evaluate our method on two benchmarks:  
LERF~\cite{kerr2023lerf} dataset comprises in-the-wild scenarios captured with the iPhone App Polycam. The objects in LERF include both common and long-tail categories with different sizes.
Scannet~\cite{dai2017scannet} is an indoor dataset comprising each annotated with instance-level segmentation and labels across 200 categories. We use 10 RGB sequences selected by~\cite{peng2023openscene} without using the depth ground truth or any human annotations.

\textbf{Implementation Details.}
\label{imple}
We implement \ourmethod{} based on HiCoM with a fixed $\lambda_1=0.25$, $\lambda_2 = 0.1$, and $\lambda_3 = 0.15$. Each incoming frame is optimized with 100 motion steps, plus another 100 steps after adding new Gaussians. 
Every fifth frame is used as a test view.
All training and testing data remain unseen to the off-the-shelf pretrained models to ensure a fair evaluation. All experiments are conducted on a single A100 80GB GPU. For more details, please refer to the Appendix.

\subsection{Quantitative and Qualitative Comparisons}
Our method enables holistic 3D object extraction across diverse tasks, including photo-realistic rendering, instance segmentation, and geometric reasoning (e.g., 3D bounding boxes, semantic occupancy, 3D mesh). We validate the effectiveness of our method through comparisons with state-of-the-art approaches and enhanced baselines in 3D reconstruction and online perception.

\textbf{Compared with reconstruction-based understanding methods.}
We compare \ourmethod{} with NeRF-based~\cite{kerr2023lerf} and Gaussian-based~\cite{qin2024langsplat, ye2024gaussian, wu2024opengaussian, qiu2024feature, li2024instancegaussian} approaches for 3D scene reconstruction with understanding. These methods rely on offline training with access to all scene views as input.  
Notably, the compared baselines also require camera poses from GT or Colmap estimated. 
For fair comparison, we incrementally replace our estimated poses with those from Colmap (denoted as \ourmethod{}$^{*}$).

Results across multiple specific tasks are presented in Table~\ref{tab:comparison}.
~\cite{ye2024gaussian, li2024instancegaussian, qin2024langsplat} utilize 2D semantic decoded by SAM as supervisions. While effective in 2D segmentation, this strategy fails to learn continuous 3D semantic-geometric representations.
Our primary competitors~\cite{qiu2024feature, wu2024opengaussian} incorporate semantic features but suffer from excessive redundant Gaussians and fail to achieve efficient joint convergence of geometry and semantics.
Moreover, all the aforementioned methods rely on complete prior observations of the 3D space, which severely limits their applicability in real-world scenes. In contrast, \ourmethod{} adopts an online training strategy that delivers high-quality reconstruction and understanding, while offering better scalability.

\begin{figure}[t] 
\centering
\includegraphics[width=1.0\textwidth]{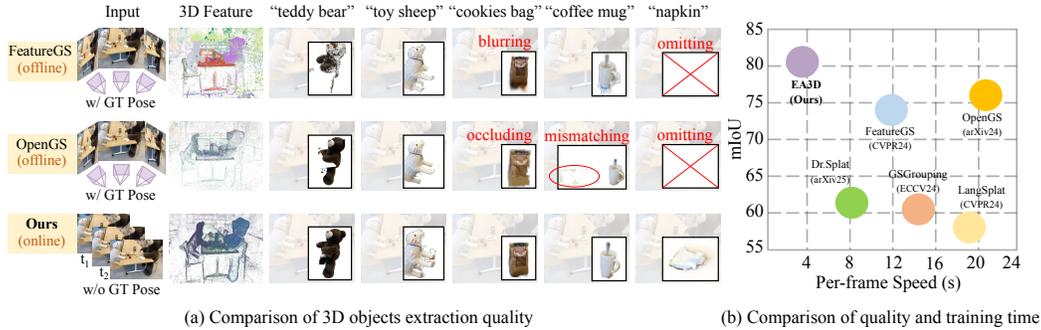}
\caption{\textbf{Visualization performance and model efficiency comparison with state-of-the-art methods.} Left (a): Under the more challenging streaming setting without pose input, \ourmethod{} delivers high-quality 3D object reconstruction and rendering. Notably, our method avoids redundant Gaussian features through efficient online updates, enabling more precise and lightweight optimization. Right (b): \ourmethod{} strikes a balance between speed and quality, significantly reducing training time while maintaining high-performance scene understanding.
}
\label{fig:speed}
\end{figure}

\setlength\tabcolsep{1.pt}
\begin{table*}
\small
\caption{Comparisons under sparse views and online incremental settings on LeRF~\cite{kerr2023lerf}. The best results are highlighted in \textbf{bold}, and the second-best results are \underline{underscored}.
``$-$'' indicates methods do not support the specified task. ``colmap'' denotes offline pose estimation using COLMAP, ``self.'' refers to online self-estimated poses. ``Speed'' denotes the average per-frame optimization speed.}
\begin{tabular*}{\linewidth}{@{\extracolsep{\fill}} lccc | ccc | ccc }
\toprule
\multicolumn{4}{c}{Tasks:} & \multicolumn{3}{c}{Rec.(PSNR $\uparrow$)} & \multicolumn{3}{c}{Seg.(mIoU $\uparrow$)} \\
\midrule
Method & Online & Pose & Speed.(FPS) & 10 views & 30 views & 70 views & 10 views & 30 views & 70 views \\ 
\midrule
LangSplat~\cite{qin2024langsplat} & \textcolor{red}{\usym{2717}} & colmap & 0.007 & 11.3  & 14.4  & 17.8  & 28.6  & 34.4  & 51.5 \\
FeatureGS~\cite{qiu2024feature} & \textcolor{red}{\usym{2717}} & colmap & 0.018 & 15.2  & 18.9  & 22.4  & 29.4  & \underline{41.2}  & 53.6 \\
OpenGaussian~\cite{wu2024opengaussian} & \textcolor{red}{\usym{2717}} & colmap & 0.005 & 14.9  & \underline{19.5}  & \underline{22.7}  & 30.1  & 40.5 & \underline{55.8} \\
\midrule
Enhanced Baselines: \\
\midrule
Cut3R~\cite{wang2025continuous}+VFM~\cite{ren2024grounded} & \textcolor{ForestGreen}{\usym{2713}} & self. & \textbf{0.648} & -  & -  & -  & 33.7  & 26.5  & 21.9 \\
HiCOM~\cite{gao2024hicom}+VFM~\cite{ren2024grounded} & \textcolor{ForestGreen}{\usym{2713}} & colmap & 0.102 & \underline{18.1}  & 18.6  & 21.5  & \underline{36.1}  & 39.3  & 43.3 \\
\midrule
\textbf{\ourmethod{}} & \textcolor{ForestGreen}{\usym{2713}} & self. & \underline{0.235} & \textbf{21.9}  & \textbf{21.8}  & \textbf{23.2}  & \textbf{53.8}  & \textbf{55.0}  & \textbf{57.4} \\
\bottomrule
\end{tabular*}
\label{tab:comparison2}
\vspace{-5pt}
\end{table*}

\textbf{Compared with online 3D scene understanding methods.} 
Two common limitations can be observed across these approaches: 1) reliance on predefined geometry or 3D representations (e.g., point clouds, depth maps, meshes); 2) dependence on extensive training with large-scale annotated datasets. As shown in Table~\ref{tab:comparison}, our method achieves competitive performance even when compared to models trained specifically for the 3D understanding tasks. ~\cite{yang2023sam3d, peng2023openscene, xu2024embodiedsam} utilize SAM to obtain 2D segmentations and project them into 3D space, but suffer from semantic ambiguities and multi-view inconsistency caused by mis-projections.
In contrast, our approach jointly optimizes geometry and knowledge without relying on 3D priors, demonstrating the strengths of our unified online framework.

\textbf{Compared with enhanced baselines.} Since our work is the first to enable online joint geometry reconstruction and scene understanding, we enhance existing methods in two ways to serve as stronger baselines: 1) augmenting online reconstruction methods with scene understanding capabilities (e.g., HiCOM+VFM, MonoGS+VFM); 2) enabling online optimization of feature Gaussians (e.g., FeatureGS+HiCOM).
Additionally, we incorporate an $L_1$ RGB loss into EmbodiedOcc~\cite{wu2024embodiedocc}, which was originally designed for online occupancy prediction. Table~\ref{tab:comparison} demonstrates that \ourmethod{} consistently outperforms our baseline HiCOM by integrating VFM-driven scene understanding. It also surpasses FeatureGS+HiCOM, which similarly employs semantic features and online updates, highlighting the effectiveness of our unified framework. Furthermore, compared to online SLAM-based methods~\cite{wu2024embodiedocc, matsuki2024gaussian}, \ourmethod{} achieves better results in both geometric reconstruction and scene interpretation.

\textbf{Qualitative Comparisons.} We further compare the visual quality of 3D object extraction with the baseline methods in Fig.~\ref{fig:speed}(a). Given a streaming video without pose information, \ourmethod{} allows high-quality reconstruction and rendering of arbitrary 3D objects. Visualizations of the 3D features show that our online feature Gaussians efficiently and accurately capture both geometry and semantics. In contrast, leading baselines introduce redundant noise, produce inferior renderings, and fail to extract challenging objects (e.g., a small piece of napkin). \ourmethod{} also enables a variety of downstream applications, such as manipulation simulation, motion emulation, controllable 3D editing, and object insertion or removal. Additional results and applications are presented in the Appendix.

Our experimental results and theoretical analyses reveal that naïve integrations of existing models tend to perform poorly and may even degrade overall performance due to inherent conflicts among components. In contrast, our method fully harnesses the open-vocabulary features extracted by VFMs and effectively tackles the key challenges of 3D semantic consistency and online geometric reconstruction. Moreover, it achieves higher efficiency and lower computational overhead through a unified and elegantly designed framework.

\subsection{Sparse Views and Online Stability} 
Table~\ref{tab:comparison2} reports the performance and robustness of \ourmethod{} under sparse-view and online incremental settings. We evaluate it by sequentially inputting sparse-view images (e.g., 10 views) and progressively extending the sequence length. In contrast, offline baselines~\cite{qin2024langsplat, qiu2024feature, wu2024opengaussian} receive all training views at once. Results show that our method exhibits strong robustness to sparse-view inputs, achieving promising results even with a few initial frames in the early stage. As the sequence length increases (10 $\rightarrow$ 30 $\rightarrow$ 70 views), \ourmethod{} maintains stable quality, while baseline methods struggle with instability and slow convergence under sparse inputs. Fig.~\ref{fig:online} further illustrates the online updating process of rendering and segmentation, occupancy estimation, and 3D mesh generation with \ourmethod{}.

\setlength\tabcolsep{16pt}
\begin{table*}[h]
    \small
    \centering
    \captionsetup{type=table}
    \caption{Ablation on key components. ``Train'' and ``Render'' represent the per-frame training and rendering time, measured in FPS. ``regular.term'' denotes the semantic-awareness regularization. ``online.opt'', ``online.odo'', and ``joint.opt'' denote the online updating strategy, online visual odometry, and joint optimization, respectively.}
    \begin{tabular}{ lcccccc }
        \toprule
        Strategy & PSNR & mIoU & mAcc & Train & Render \\
        \midrule
        Baseline: HiCoM~\cite{gao2024hicom} & 22.6  & 34.8  & 61.9  & 0.29 & 230 \\\
        W/o CLIP Encoder & 25.3  & 41.6  & 66.4  & 0.28 & 220 \\
        W/o SAM Encoder & 25.4  & 42.8  & 67.1  & 0.27  & 215 \\
        W/o regular.term & 25.1  & 44.3  & 70.5  & 0.21 & 208 \\
        W/o online.opt & 24.6  & 44.5  & 69.7  & 0.07  & 110 \\
        W/o online.odo & 25.0  & 45.4  & 70.8  & 0.26  & 205 \\
        W/o joint.opt & 24.8  & 45.7  & 71.4  & 0.25 & 210 \\
        \textbf{Ours-full} & 25.8  & 46.3  & 71.8 & 0.23  & 210 \\
        \bottomrule
    \end{tabular}
    \label{tab:ablation}
    \vspace{-7pt}
\end{table*}

\subsection{Model Efficiency Analysis}
Our method enables online incremental reconstruction and understanding of scenes for 3D object extraction. Here, we quantitatively evaluate the speed and memory usage of each key component. As shown in Fig.~\ref{fig:speed}(b), our method achieves faster optimization while maintaining top performance. \ourmethod{} strikes a balance between speed and accuracy, delivering higher rendering efficiency with reduced storage overhead. Detailed quantitative experimental results are provided in the Appendix.

\subsection{Ablation Studies}
\label{sec:ablation}
As shown in Table~\ref{tab:ablation}, we conduct ablation studies and analyze the key components of our designs for online open-world 3D object extraction. 
Embedded visual features from VFMs (e.g., CLIP~\cite{gao2024clip} and SAM~\cite{ravi2024sam}) imbue Gaussians with semantic awareness, enhancing both fine-grained geometry modeling and scene understanding. Our online optimization strategy accelerates feature Gaussian refinement via an efficient feedforward mechanism, ensuring accuracy while minimizing redundancy.
The online visual odometry provides dynamic pose updates and dense geometric cues, speeding up convergence. 
Semantic-aware regularization links Gaussian geometry with semantic features, ensuring object-level 3D consistency and smoothness. 
By jointly optimizing geometry, semantics, and pose, our method enables recurrent feature updates that seamlessly integrate appearance and structure for robust 3D reconstruction and understanding. For more ablation studies on key modules and hyperparameters, please refer to the Appendix.

\section{Conclusion}\label{sec:con}
We have presented \ourmethod{}, a unified online framework for open-world 3D object extraction.
\ourmethod{} enables simultaneous online reconstruction and understanding without geometric or pose priors. 
It consistently achieves good performance across a broad set of tasks, including photo-realistic reconstruction and rendering, semantic and instance segmentation, 3D bounding box construction, semantic occupancy estimation, and 3D mesh generation. \ourmethod{} introduces a novel perspective for aligning and aggregating 3D semantic and geometric features through online reconstruction and dynamic update strategies. It establishes a unified online 3D feature aggregation framework grounded in reconstruction constraints, enabling more accurate and efficient 3D scene understanding and reconstruction.
\section*{Acknowledgment}
This work was supported by National Key R\&D Program of China (Grant No. 2022ZD0160305). This work was also a research achievement of Key Laboratory of Science, Technology, and Standard in Press Industry (Key Laboratory of Intelligent Press Media Technology). Ming-Hsuan Yang was supported in part by the Institute of Information \& Communications Technology Planning \& Evaluation (IITP) grant funded by the Korean Government (MSIT) (No. RS-2024-00457882, National AI Research Lab Project).

\section*{Broader Impacts}
\label{impact}
This paper presents research aimed at advancing the fields of 3D vision, which hold significant promise for enhancing the 3D object extraction. While AI-driven scene reconstruction and perception bring benefits, they could also raise concerns regarding their social and economic impacts. Automating 3D labeling and perception tasks can potentially disrupt the labor market, posing risks to certain job sectors, particularly in sectors that rely on manual data annotation. It is crucial to exercise caution and ensure that the societal implications are thoroughly addressed.

{\small
\bibliographystyle{plain}
\bibliography{main}
}

\newpage

\section*{NeurIPS Paper Checklist}

\begin{enumerate}

\item {\bf Claims}
    \item[] Question: Do the main claims made in the abstract and introduction accurately reflect the paper's contributions and scope?
    \item[] Answer: \answerYes{} 
    \item[] Justification: We claim the main contribution of this paper in both the Abstract and Introduction sections.

\item {\bf Limitations}
    \item[] Question: Does the paper discuss the limitations of the work performed by the authors?
    \item[] Answer: \answerYes{} 
    \item[] Justification: We discuss the limitation of this work in the Supplementary materials.

\item {\bf Theory assumptions and proofs}
    \item[] Question: For each theoretical result, does the paper provide the full set of assumptions and a complete (and correct) proof?
    \item[] Answer: \answerNA{} 
    \item[] Justification: This paper does not include theoretical results.

    \item {\bf Experimental result reproducibility}
    \item[] Question: Does the paper fully disclose all the information needed to reproduce the main experimental results of the paper to the extent that it affects the main claims and/or conclusions of the paper (regardless of whether the code and data are provided or not)?
    \item[] Answer: \answerYes{} 
    \item[] Justification: We provide the implementation details in Section \ref{imple}.

\item {\bf Open access to data and code}
    \item[] Question: Does the paper provide open access to the data and code, with sufficient instructions to faithfully reproduce the main experimental results, as described in supplemental material?
    \item[] Answer: \answerNo{} 
    \item[] Justification: We do not provide new datasets and will release partial code after the paper is accepted.

\item {\bf Experimental setting/details}
    \item[] Question: Does the paper specify all the training and test details (e.g., data splits, hyperparameters, how they were chosen, type of optimizer, etc.) necessary to understand the results?
    \item[] Answer: \answerYes{} 
    \item[] Justification:  We provide the training details and hyperparameters in Section \ref{imple}.

\item {\bf Experiment statistical significance}
    \item[] Question: Does the paper report error bars suitably and correctly defined or other appropriate information about the statistical significance of the experiments?
    \item[] Answer: \answerNo{} 
    \item[] Justification: Error bars are not reported because it would be too computationally expensive.

\item {\bf Experiments compute resources}
    \item[] Question: For each experiment, does the paper provide sufficient information on the computer resources (type of compute workers, memory, time of execution) needed to reproduce the experiments?
    \item[] Answer: \answerYes{} 
    \item[] Justification: We provide the information for computer resources in Section \ref{imple}.
    
\item {\bf Code of ethics}
    \item[] Question: Does the research conducted in the paper conform, in every respect, with the NeurIPS Code of Ethics \url{https://neurips.cc/public/EthicsGuidelines}?
    \item[] Answer: \answerYes{} 
    \item[] Justification: The research in the paper conforms with the NeurIPS Code of Ethics.

\item {\bf Broader impacts}
    \item[] Question: Does the paper discuss both potential positive societal impacts and negative societal impacts of the work performed?
    \item[] Answer: \answerYes{} 
    \item[] Justification: We provide the discussion of broader impacts in the Appendix.
    
\item {\bf Safeguards}
    \item[] Question: Does the paper describe safeguards that have been put in place for responsible release of data or models that have a high risk for misuse (e.g., pretrained language models, image generators, or scraped datasets)?
    \item[] Answer: \answerNA{} 
    \item[] Justification: The models in this paper pose no such risks.

\item {\bf Licenses for existing assets}
    \item[] Question: Are the creators or original owners of assets (e.g., code, data, models), used in the paper, properly credited and are the license and terms of use explicitly mentioned and properly respected?
    \item[] Answer: \answerYes{} 
    \item[] Justification: All owners of models, code, and data we used are properly cited. We compliance all licenses of models, code, and data.

\item {\bf New assets}
    \item[] Question: Are new assets introduced in the paper well documented and is the documentation provided alongside the assets?
    \item[] Answer: \answerNA{} 
    \item[] Justification: The paper does not release new assets.

\item {\bf Crowdsourcing and research with human subjects}
    \item[] Question: For crowdsourcing experiments and research with human subjects, does the paper include the full text of instructions given to participants and screenshots, if applicable, as well as details about compensation (if any)? 
    \item[] Answer: \answerNA{} 
    \item[] Justification: This paper does not involve crowdsourcing or research with human subjects.

\item {\bf Institutional review board (IRB) approvals or equivalent for research with human subjects}
    \item[] Question: Does the paper describe potential risks incurred by study participants, whether such risks were disclosed to the subjects, and whether Institutional Review Board (IRB) approvals (or an equivalent approval/review based on the requirements of your country or institution) were obtained?
    \item[] Answer: \answerNA{} 
    \item[] Justification: This paper does not involve crowdsourcing or research with human subjects.

\item {\bf Declaration of LLM usage}
    \item[] Question: Does the paper describe the usage of LLMs if it is an important, original, or non-standard component of the core methods in this research? Note that if the LLM is used only for writing, editing, or formatting purposes and does not impact the core methodology, scientific rigorousness, or originality of the research, declaration is not required.
    \item[] Answer: \answerYes{} 
    \item[] Justification: We describe the usage of LLMs in Section~\ref{sec:exp}.

\end{enumerate}

\clearpage
\section*{Appendix}\appendix
\label{appendix}

\newcommand{\applabel}{Appendix\xspace}
\renewcommand{\thesection}{\Alph{section}}
\renewcommand{\thetable}{\Roman{table}}
\renewcommand{\thefigure}{\Roman{figure}}
\setcounter{section}{0}
\setcounter{table}{0}
\setcounter{figure}{0}

In this appendix, we provide additional content to complement the main paper:
\begin{itemize}[leftmargin=1em,topsep=0pt]

\item \applabel~\ref{data}: Datasets and Implementation Details.

\item \applabel~\ref{method}: Method Details.

\item \applabel~\ref{enhancing}: Details of enhancing and comparing with our baseline methods.

\item \applabel~\ref{novelty}: Novelty clarification against baselines.

\item \applabel~\ref{effi}: Model Efficiency Analysis.

\item \applabel~\ref{ablations}: Detailed Ablation Studies.

\item \applabel~\ref{visual}: More Qualitative Visualizations.

\item \applabel~\ref{application}: Diverse Downstream Applications.

\item \applabel~\ref{failure}: Failure Cases and Limitations.

\item \applabel~\ref{impact2}: Broader Impacts.

\end{itemize}

\section{Datasets and Implementation Details}
\label{data}

\textbf{Datasets.} We evaluate our method on two benchmarks. The LERF~\cite{kerr2023lerf} dataset, captured with the iPhone Polycam app, features complex in-the-wild scenes. We use the extended version from~\cite{qin2024langsplat}, which includes ground-truth annotations for 3D object localization and 3D semantic segmentation. In addition, we manually annotated challenging open-vocabulary categories and hard cases to enable a more comprehensive evaluation of our method. The ScanNet~\cite{dai2017scannet} dataset comprises a diverse set of indoor scenes with a rich variety of objects. While it offers RGB-D images and 3D meshes, our pipeline utilizes only the RGB image sequences. Consistent with prior work such as EmbodiedSAM~\cite{xu2024embodiedsam}, we use the same high-quality indoor scenes and labeled point clouds for evaluation.
For semantic evaluation, we compute metrics using all ground truth classes from LERF and ScanNet. 
Categories predicted by the VLMs may be absent from the ground truth due to the benchmark’s limited semantic classes. To ensure consistency with baseline methods, we prompt VLMs to merge such categories with the closest predefined classes—for example, combining “bookshelf” and “bookcase” under “bookcase.”

\textbf{Implementation Details.} 
We implement \ourmethod{} on top of HiCoM, with reduced training iterations to ensure rapid Gaussian updates. Each incoming frame undergoes 100 update steps, followed by another 100 training steps after the incorporation of new Gaussians. The Gaussian parameters are initially initialized based on the estimated odometry and corresponding camera poses. Low-opacity Gaussians are removed prior to training on the next frame, allowing them to still contribute to the current representation. We employ the off-the-shelf CogVLM~\cite{wang2024cogvlm, hong2024cogvlm2} model to interpret the scene. For semantic feature map extraction, we utilize the Grounded-SAM and CLIP models, with ViT-Huge serving as the image encoder. To enhance efficiency, we apply a 2× downsampling to the input image before feeding it into the feature extractor. At the time of this work, the official code released by baseline methods exhibited instability and execution issues. Therefore, we report experimental results based on our own implementation. All experiments are conducted using PyTorch on a single 80GB A100 GPU.

\section{Method Details}
\label{method}

\begin{figure}[t] 
\centering
\includegraphics[width=1.0\textwidth]{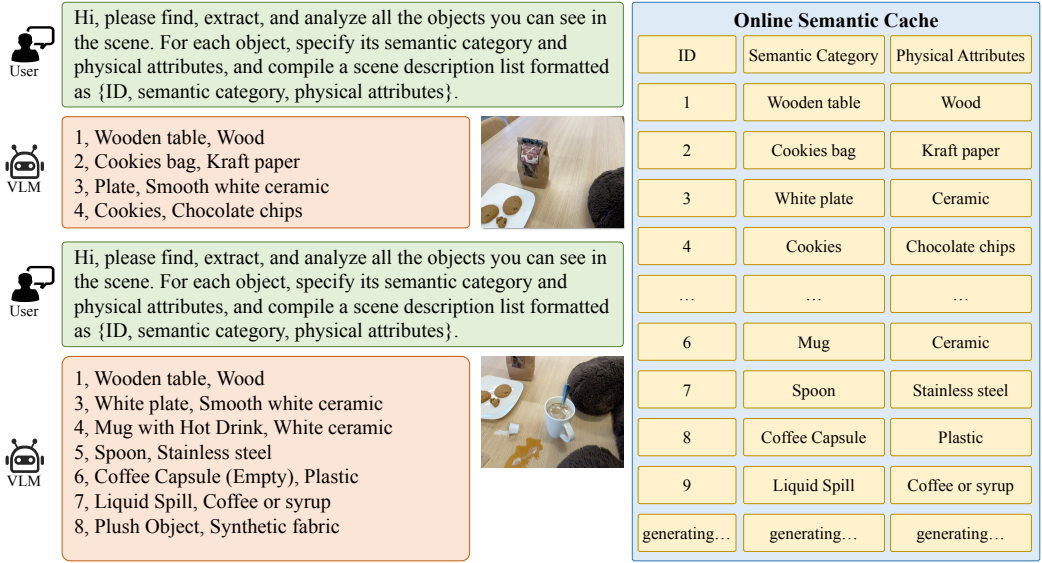}
\caption{\textbf{Visualization of Open-world interpretation by VLMs.} 
}
\label{fig:vlm}
\vspace{-15pt}
\end{figure}

\textbf{Open-world interpretation by VLMs.}
We present a detailed illustration of how open-world scene understanding is obtained online from VLMs, as shown in Fig.~\ref{fig:vlm}. We use the key prompt ``find, identify, and analyze anything in the scene'' to guide VLMs in extracting object categories from single-frame images, which are then dynamically updated into an online semantic cache. Notably, the semantics extracted by VLMs may contain ambiguities or redundancies. We address semantic ambiguities in the Method section of the main text. To reduce redundancy, we adopt a semantic fusion strategy that avoids repeatedly storing similar or overlapping concepts in the cache. Specifically, each semantic label is encoded into a feature vector $ T \in \mathbb{R}^{1 \times V} $ using a pretrained text encoder from CLIP~\cite{Zhou_2022_CVPR, yu2023turning}. We compute pairwise similarities between these vectors and merge those exceeding a predefined similarity threshold $\vartheta$. For example, “brown toy bear” and “brown teddy bear” are merged, while semantically distinct concepts like “chair” and “sofa” remain separate. More ablation about the semantic cache updating threshold $\vartheta$ is further conducted in Section~\ref{ablations}. For semantic cache updating threshold, we first employ an aggregation strategy for physical attributes via instance-level feature map fusion, performed during the online cache update. In this process, physical attribute features with the highest occurrence frequency and confidence are dynamically fused into the online semantic cache as a variable-length vector, under the constraint of multi-view 3D consistency.

\textbf{Online Gaussian Splatting.} 3D Gaussian Splatting explicitly represents scenes using anisotropic 3D Gaussian primitives, including position $\mu$, covariance matrix $\Sigma$, opacity $o$, and spherical harmonics coefficients (SH):
\begin{equation}
\label{formula:gaussian's formula}
    G(\mathbf{x})=e^{-\frac{1}{2}(\mathbf{x} - \boldsymbol{\mu})^T\mathbf{\Sigma}^{-1}(\mathbf{x} - \boldsymbol{\mu})}.
\end{equation}
The covariance matrix $\mathbf{\Sigma}$ is decomposed into a scaling matrix $\mathbf{S}$ and a rotation matrix $\mathbf{R}$ to ensure physical meaning and facilitate optimization:
\begin{equation}
\label{formula:covariance decomposition}
    \mathbf{\Sigma} = \mathbf{R}\mathbf{S}\mathbf{S}^T\mathbf{R}^T,
\end{equation}
where $\mathbf{S} = \text{diag}(s_x, s_y, s_z) \in \mathbb{R}^3$ and $\mathbf{R} \in SO(3)$ are parameterized by a 3D scaling vector $\mathbf{s}$ and a rotation quaternion $\mathbf{q}$, respectively. Each Gaussian primitive is further enriched with color and opacity, represented by spherical harmonic coefficients $\mathbf{h}$ and a scalar $\alpha$, respectively.
We further augment GS with fused features from VLMs and VFMs, comprising semantic features $\mathbf{S}$, physical attribute features $\mathbf{Y}$, and a continuous vector $ T \in \mathbb{R}^{1 \times V} $ retrieved from an online semantic cache $\Omega$. To render a novel viewpoint, Gaussian primitives are projected onto the camera plane with alpha-blending to accumulate the final splatted feature $\hat{F}$: 
\begin{align}
\hat{F} = \sum_{i \in N} F_i \cdot \alpha_i \prod_{j=1}^{i-1}(1-\alpha_j) , 
\end{align}
where $\alpha_i$ denotes the opacity, $F_i$ is the integrated feature map of the $i$-th Gaussian. The contributions of $N$ overlapping Gaussian primitives at each pixel account for their depth-ordering.

\textbf{Gaussian2Voxel Splatting.} 
Inspired by~\cite{zhou2025autoocc}, we use accumulated Gaussians to splat onto the voxel grid at an arbitrary voxel size to generate the occupancy, with each voxel’s occupancy determined by weighting the occupied range and opacity of the Gaussians:
\begin{equation}
\begin{aligned}
    \digamma(o) = \sum_{i=1}^{N} d_i G(x_i) \alpha_{i} \mathrm{softmax}(\mathbf{F}_t) ,
\end{aligned}
\end{equation}
where $d_i$ is the occupied depth of the Gaussian2voxel, treated as the splatting weight coefficient. $\alpha_{i}$ is the opacity, $\mathbf{F}_t$ is the integrated feature map.

\textbf{3D Bbox Estimation.} For each online feature Gaussian, we generate category-specific boundaries by applying a KNN clustering algorithm to select the boundary ranges of Gaussian ellipsoids sharing the same semantic category. The spatial coordinates of semantic cluster centers serve as the 3D bounding box centers. The bounding box dimensions (i.e., length, width, and height) are determined by applying the Axis-Aligned Bounding Box (AABB) algorithm to enclose the Gaussian ellipsoids within the cluster, based on their intersections with the bounding box edges.

\begin{figure}[t] 
\centering
\includegraphics[width=1.0\textwidth]{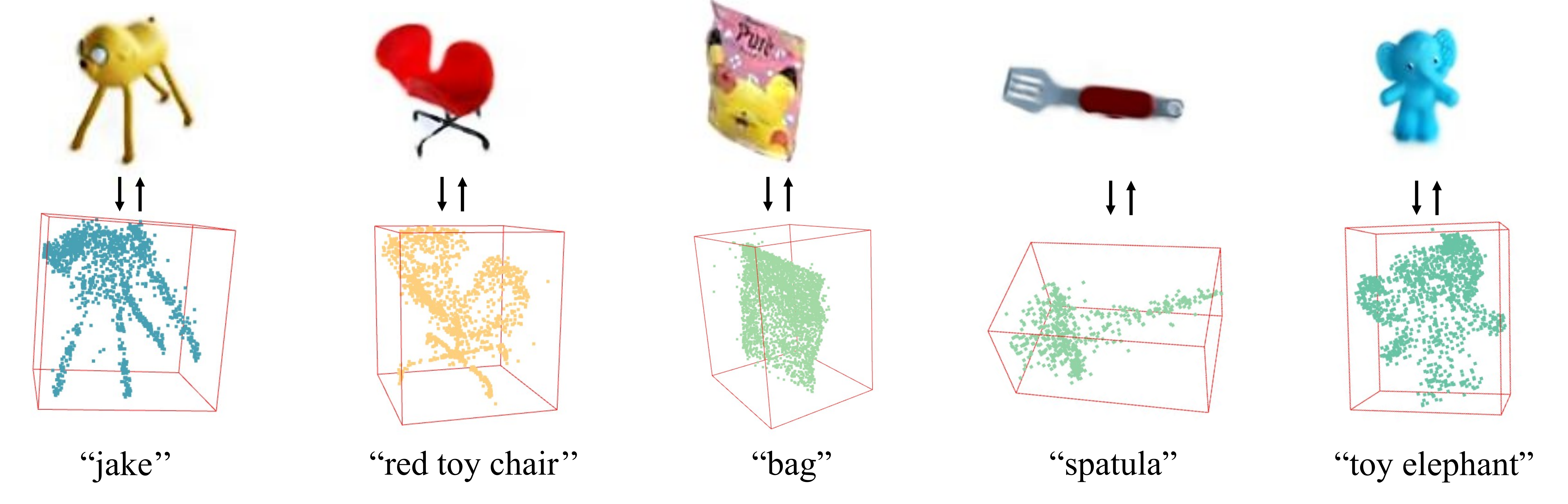}
\caption{\textbf{Visualization of Semantic-aware splatting to 3D Bbox and Semantic Occupancy.} 
}
\label{fig:bbox-vis}
\vspace{-10pt}
\end{figure}

\textbf{3D Mesh Generation.} Following PGSR~\cite{chen2024pgsr}, we start surface extraction by rendering the depth from our online feature Gaussians for each training view. We then apply the TSDF Fusion~\cite{newcombe2011kinectfusion, wei2024gsfusion} algorithm to construct the corresponding TSDF field, from which the mesh is subsequently extracted.

\section{Details of Enhancing and Comparing with Baseline Methods}
\label{enhancing}
Since we are the first to propose online 3D object extraction without relying on 3D geometric priors, poses and predefined category lists, we enhance prior and concurrent methods to serve as stronger baselines. All enhanced baselines are reimplemented and refactored from their original codebases. Detailed implementation will be made available upon acceptance of the paper.

\textbf{Streaming Gaussian + Open-Vocabulary Segmentation.}  
Although streaming Gaussian-based methods~\cite{sun20243dgstream, gao2024hicom, yan2025instant} enable scene reconstruction from video streams, they suffer from critical limitations, including the need for initial multi-view coverage and pre-known camera poses. Moreover, their inability to understand the scene semantics makes them unsuitable for the 3D object extraction task. To address this issue, we enhance our main baseline HiCOM~\cite{gao2024hicom} by integrating online streaming Gaussian optimization with VFM-guided semantics. Specifically, HiCOM incrementally reconstructs the scene Gaussians frame by frame, while each frame's 2D semantic segmentation—generated by VFMs—is lifted and projected onto the reconstructed Gaussians through a two-stage 2D-to-3D mapping process. As shown in Tab.~\ref{tab:comparison3}, our method outperforms the enhanced baseline, achieving a better trade-off between accuracy and speed. In contrast, the compared baselines exhibit noticeable quality degradation due to the lack of joint optimization for scene geometry and understanding.

\textbf{Online SLAM + Open-Vocabulary Segmentation.}
SLAM-based methods allow for online mapping of scenes with unknown poses but are highly dependent on accurate geometric priors from depth input and expensive post-refinement. They also fail to simultaneously reconstruct geometry and understand scenes. To overcome this challenge, we integrate VFM-guided semantics into SLAM-based online mapping systems, synchronously projecting the acquired semantic priors onto the constructed point cloud or 3D Gaussians. For a fair comparison, the baseline excludes the post-refinements, which are typically considered offline procedures. As shown in Table~\ref{tab:comparison3}, SLAM-based methods struggle to jointly recover accurate geometry and semantics without costly post-processing and face ambiguity in complex scenes~\cite{kerr2023lerf}.

\textbf{Feature-distillation Gaussian + Streaming Update.}
Feature Gaussians~\cite{qiu2024feature, zhou2025feature4x, zhou2024feature} propose to distill object-centric vision-language features into 3D Gaussians within the same optimization pipeline as vanilla 3DGS. However, these methods operate offline and cannot support online 3D object extraction. We introduce an online Gaussian update~\cite{gao2024hicom} combined with feature distillation~\cite{qiu2024feature} to achieve semantic-aware online Gaussians. Notably, since the feature-carrying 3D Gaussians proposed by~\cite{qiu2024feature, zhou2024feature} do not support incremental online updates, we first use HiCOM to add new Gaussians per frame, then distill features into them sequentially. While this two-stage process largely preserves the effectiveness of the original method, it introduces significant runtime disruptions. As shown in Table~\ref{tab:comparison3}, our method outperforms this strong baseline by enabling end-to-end online updating and joint optimization of feature-rich Gaussians, enhancing performance while maintaining model efficiency.

\textbf{Cut3R + SAM3D.}
Cut3R~\cite{wang2025continuous} enables the online generation of metric-scale point maps (per-pixel 3D points) from video streams. However, Cut3r struggles to preserve fine geometric details, photorealistic rendering, and scene understanding. It can also generate extremely blurry and distorted results when extrapolating far from observed views. We enhance Cut3r with scene understanding by integrating SAM3D~\cite{yang2023sam3d}, employing a bidirectional merging strategy to project 2D masks into 3D. Results in Table~\ref{tab:comparison3} show that our method outperforms the extended Cut3r, delivering higher-quality geometry and rendering without a significant increase in computational cost. Moreover, Cut3r generates a large number of redundant per-pixel 3D points, which interfere with semantic projection. In contrast, our method employs an online Gaussian update strategy to remove redundant Gaussians while implicitly aligning semantics in 3D space.

\setlength\tabcolsep{1.5pt}
\begin{table*}
\small
\caption{Comparisons on ScanNet~\cite{dai2017scannet}. The best results are highlighted in \textbf{bold}, and the second-best results are \underline{underscored}.
``$*$'' indicates the use of the colmap-estimated poses following~\cite{ye2024gaussian, qin2024langsplat, qiu2024feature}. ``$-$'' indicates that the method does not support the specified task. ``Rec., Seg., Bbox., Occ.'' denotes four multi-task evaluations: reconstruction quality, instance segmentation, 3D bounding box estimation, and semantic occupancy estimation. ``Speed'' refers to the training speed, measured in frames per second (FPS).
}
\begin{tabular*}{\linewidth}{@{\extracolsep{\fill}} lccc | cc | cc | cc | cc}
\toprule
\multicolumn{4}{c}{Task:} & \multicolumn{2}{c}{Rec.} & \multicolumn{2}{c}{Seg.} & \multicolumn{2}{c}{Bbox.} & \multicolumn{2}{c}{Occ.} \\
\midrule
Method & Input & Online & Speed & PSNR & SSIM & mIoU & mAcc & AP & mAP & IoU & mIoU \\ 
\midrule
HiCOM~\cite{gao2024hicom} & RGB & \textcolor{ForestGreen}{\usym{2713}} & 0.29 & 22.6  & 0.82  & 34.8 & 61.9  & \textcolor{red}{\usym{2717}}  & \textcolor{red}{\usym{2717}}  & \textcolor{red}{\usym{2717}}  & \textcolor{red}{\usym{2717}} \\
HiCOM~\cite{gao2024hicom}+VFM~\cite{ren2024grounded} & RGB & \textcolor{ForestGreen}{\usym{2713}} & 0.11 & 22.6  & 0.82  & 34.8  & 61.9 & 52.5  & 23.8  &  42.4  & 27.9 \\
\midrule
MonoGS~\cite{matsuki2024gaussian} & RGBD & \textcolor{ForestGreen}{\usym{2713}} & 0.18 & 24.3  & 0.85  & 36.3 & 60.5  & \textcolor{red}{\usym{2717}}  & \textcolor{red}{\usym{2717}}  & \textcolor{red}{\usym{2717}}  & \textcolor{red}{\usym{2717}} \\
MonoGS~\cite{matsuki2024gaussian}+VFM~\cite{ren2024grounded} & RGBD & \textcolor{ForestGreen}{\usym{2713}} & 0.07 & 24.3  & 0.85  & 36.3  & 60.5 & 51.7 & 27.7 &  44.5  & 27.2 \\
SGS-slam~\cite{li2024sgs}+VFM~\cite{ren2024grounded} & RGBD & \textcolor{ForestGreen}{\usym{2713}} & 0.05 & 20.7  & 0.78  & 33.5 & 57.8  & 45.6  & 25.2  & 35.4  & 22.0 \\
\midrule
FeatureGS~\cite{qiu2024feature} & RGB & \textcolor{red}{\usym{2717}} & 0.01 & 23.9  & 0.84  & 41.1  & 66.0  & 51.4  & 32.7  & 50.9  & 31.2 \\
FeatureGS~\cite{qiu2024feature}+HiCOM~\cite{matsuki2024gaussian} & RGB & \textcolor{ForestGreen}{\usym{2713}} & 0.03 & 24.5  & 0.85  & 40.8  & 66.3 & 55.8  & 34.7  &  50.7  & 31.4 \\
Feat-3dgs~\cite{zhou2024feature}+HiCOM~\cite{matsuki2024gaussian} & RGB & \textcolor{ForestGreen}{\usym{2713}} & 0.02 & 23.3  & 0.84  & 38.9 & 63.5  & 50.1  & 28.6  & 49.2  & 30.5\\
\midrule
LSM~\cite{fan2024large} & RGB  & \textcolor{ForestGreen}{\usym{2713}} & 0.89 & 24.3  & 0.80  & 40.2  & 61.7  & \textcolor{red}{\usym{2717}}  & \textcolor{red}{\usym{2717}} & \textcolor{red}{\usym{2717}} & \textcolor{red}{\usym{2717}} \\
SAM3D~\cite{yang2023sam3d} & Points & \textcolor{ForestGreen}{\usym{2713}} & 0.92 & \textcolor{red}{\usym{2717}}  & \textcolor{red}{\usym{2717}}  & 39.2  & 62.3  & 53.7  & 29.1  & 53.3  & 26.7 \\
Cut3R~\cite{wang2025continuous}+SAM3D~\cite{yang2023sam3d} & RGB & \textcolor{ForestGreen}{\usym{2713}} & 0.41 & \textcolor{red}{\usym{2717}}  & \textcolor{red}{\usym{2717}}  & 40.3  & 62.5 & 50.6  & 26.4  &  46.6  & 25.3 \\
\midrule
\textbf{\ourmethod{}$^*$} & RGB & \textcolor{ForestGreen}{\usym{2713}} & 0.20 & \underline{25.5}  & \underline{0.87}  & \underline{45.9}  & \underline{71.2}  & \textbf{59.2}  & \underline{39.6}  & 55.0  & \textbf{34.3} \\
\textbf{\ourmethod{}} & RGB & \textcolor{ForestGreen}{\usym{2713}} & 0.23 & \textbf{25.8}  & \textbf{0.89}  & \textbf{46.3}  & \textbf{71.8}  & 57.9  & \textbf{39.9}  & \textbf{55.4}  & \underline{33.9} \\
\bottomrule
\end{tabular*}
\label{tab:comparison3}
\vspace{-5pt}
\end{table*}

\section{Novelty Clarification Against Baselines}
\label{novelty}
Here, we further clarify the distinctions and advantages of our proposed method compared to concurrent works. 

\textbf{Compared with Feature Gaussian methods.} 
Current feature Gaussian splatting methods~\cite{qiu2024feature, zhou2024feature, zhou2025feature4x, li2024langsurf} aim to equip GS with scene understanding via feature field distillation but remain tied to the fully offline vanilla 3DGS pipeline. While these approaches combine semantic feature gradients with Gaussian attribute updates, they lack an explicit joint optimization strategy for geometry and semantics, often resulting in slow convergence. Notably, the distilled features in these methods are predefined, with fixed semantic categories that remain unchanged throughout the optimization process.
In contrast, our method adopts a fully online feature embedding strategy with a simple yet effective feedforward update mechanism, enabling dynamic and adaptive feature refinement, which enhances the pipeline's generalization.

\textbf{Compared with Streaming Gaussian methods.} 
Current streaming Gaussian methods~\cite{sun20243dgstream, gao2024hicom, yan2025instant} face three challenges: 1) they require multi-view video streams, which incur high capture costs in practical applications; 2) they depend on pre-known camera poses; 3) they cannot jointly optimize scene geometry and understanding in a synchronized manner. In contrast, building on streaming Gaussians, we introduce an online visual odometry that enables incremental reconstruction from monocular dynamic video streams. Additionally, we design a knowledge-fusion streaming feature update strategy to ensure rapid optimization of both geometry and scene understanding.

\textbf{Compared with 2D-to-3D Lifting methods.}
A straightforward way to obtain 3D scene understanding from 2D foundational models is to lift the 2D results into 3D using a voting fusion algorithm combined with 2D-to-3D projection. Previous approaches~\cite{yang2023sam3d, zhang2023sam3d, nguyen2024open3dis, huang2024segment3d, chacko2025lifting} leverage SAM to segment 2D images and project the results onto pre-constructed 3D representations such as point clouds, meshes, or 3DGS. Our method differs from these approaches in two key aspects: 1) \ourmethod{} does not rely on prebuilt 3D representations but simultaneously construct scene geometry and semantics; 2) \ourmethod{} achieves efficient 3D spatial alignment through hybrid feature embeddings rather than directly projecting decoded 2D outputs. Experimental results demonstrate that our approach outperforms lifting methods, effectively addressing semantic ambiguity and occlusion issues inherent in 2D-to-3D.

\textbf{Compared with online SLAM methods.}
SLAM-based methods online 3D scene mapping without known camera poses. Recent advances~\cite{li2024sgs, li2024dns, zhu2024sni, zhu2024semgauss} extend SLAM to scene understanding by incorporating semantic information to provide additional supervision for semantic scene mapping. However, these methods require additional depth ground truth as input, which is difficult to obtain in real-world applications. They also rely on 2D semantic segmentation masks and costly post-processing for global semantic bundle adjustment. In contrast, \ourmethod{} is a fully end-to-end online 3D object extraction method that requires no geometric priors or costly post-processing. Our method offers greater flexibility, supporting open-world 3D semantic understanding and multi-level geometric construction. \ourmethod{} also outperforms these SLAM-based methods in training and rendering speed, reconstruction quality, and semantic accuracy.

\section{Model Efficiency Analysis}
\label{effi}
A comprehensive comparison of model efficiency is shown in Tab.~\ref{tab:speed}, including module-wise breakdown, training time, rendering speed and quality, model size, and memory usage. \ourmethod{} utilizes joint online visual odometry and Gaussian optimization, both of which are faster than offline approaches. Despite leveraging additional visual base models to enhance the understanding of long-tail objects in the open world, our method maintains a comparable or even faster feature embedding speed as we extract image features without the need for a decoding process. In contrast, LangSplat~\cite{qin2024langsplat} and GSGrouping~\cite{ye2024gaussian} require feature decoding during the training phase, which is time-consuming. Our method strikes a balance between speed and accuracy, ensuring higher rendering efficiency and lower storage overhead.

\begin{table*}[ht]
\centering
\setlength{\tabcolsep}{1.0mm}
\caption{\textbf{Efficiency and Performance Comparison.} Since all baseline methods perform offline reconstruction, we report the average runtime per component and total pipeline by measuring the execution time across all training views of the entire scene for them. ``Total'' indicates the average training speed per frame, ``Render'' refers to the rendering speed, and ``Parameters'' represent all trainable parameters in the pipeline.
}
\begin{tabular}{llcccccc}
\toprule
\multirow{2}{*}{Method} & \multirow{2}{*}{Component} & \multirow{2}{*}{Online} & \multirow{2}{*}{\shortstack{Component\\(FPS$\uparrow$)}} & \multirow{2}{*}{\shortstack{Total\\(FPS$\uparrow$)}} & \multirow{2}{*}{\shortstack{Quality\\(PSNR$\uparrow$)}} & \multirow{2}{*}{\shortstack{Render\\(FPS$\uparrow$)}} & \multirow{2}{*}{\shortstack{Parameters\\(M$\downarrow$)}}\\
& & & & & \\
\midrule
\multirow{3}{*}{LERF~\cite{kerr2023lerf}} & Colmap & \textcolor{red}{\usym{2717}} & 0.49 & \multirow{3}{*}{0.03} & \multirow{3}{*}{16.5} & \multirow{3}{*}{67} & \multirow{3}{*}{1272}\\
& Whole Seg. & \textcolor{red}{\usym{2717}}  & 0.12 & & & & \\
& GS Training & \textcolor{red}{\usym{2717}} & 0.06 & & & &\\
\midrule
\multirow{3}{*}{LangSplat~\cite{qin2024langsplat}} & Colmap & \textcolor{red}{\usym{2717}} & 0.49 & \multirow{3}{*}{0.08} & \multirow{3}{*}{18.4} & \multirow{3}{*}{140} & \multirow{3}{*}{714}\\
& Whole Seg. & \textcolor{red}{\usym{2717}}  & 0.13 & & & & \\
& GS Training & \textcolor{red}{\usym{2717}} & 0.26 & & & &\\
\midrule
\multirow{3}{*}{GSGrouping~\cite{ye2024gaussian}} & Colmap & \textcolor{red}{\usym{2717}} & 0.49 & \multirow{3}{*}{0.06} & \multirow{3}{*}{19.6} & \multirow{3}{*}{180} & \multirow{3}{*}{460}\\
& Whole Seg. & \textcolor{red}{\usym{2717}}  & 0.19 & & & & \\
& GS Training & \textcolor{red}{\usym{2717}} & 0.13 & & & &\\
\midrule
\multirow{3}{*}{OpenGaussian~\cite{wu2024opengaussian}} & Colmap & \textcolor{red}{\usym{2717}} & 0.49 & \multirow{3}{*}{0.05} & \multirow{3}{*}{22.1} & \multirow{3}{*}{120} & \multirow{3}{*}{528}\\
& Whole Seg. & \textcolor{red}{\usym{2717}}  & 0.14 & & & & \\
& GS Training & \textcolor{red}{\usym{2717}} & 0.10 & & & &\\
\midrule
\multirow{3}{*}{FeatureGS~\cite{qiu2024feature}} & Colmap & \textcolor{red}{\usym{2717}} & 0.49 & \multirow{3}{*}{0.07} & \multirow{3}{*}{23.9} & \multirow{3}{*}{190} & \multirow{3}{*}{647}\\
& Whole Seg. & \textcolor{red}{\usym{2717}}  & 0.23 & & & & \\
& GS Training & \textcolor{red}{\usym{2717}} & 0.15 & & & &\\
\midrule
\multirow{3}{*}{\textbf{Ours}} & Online Odo & \textcolor{ForestGreen}{\usym{2713}} & 1.67 & \multirow{3}{*}{\textbf{0.23}} & \multirow{3}{*}{\textbf{25.8}} & \multirow{3}{*}{\textbf{210}} & \multirow{3}{*}{\textbf{364}}\\
& Feature Embed. & \textcolor{ForestGreen}{\usym{2713}}  & 0.43 & & & & \\
& GS Training & \textcolor{ForestGreen}{\usym{2713}} & 0.84 & & & &\\
\bottomrule
\end{tabular} 
\label{tab:speed}
\end{table*}

\section{Detailed Ablation Studies}
\label{ablations}

\textbf{Sensitivity to the online semantic cache.}
The online semantic cache dynamically updates the extracted object categories as new observations arrive. We conduct ablation studies to evaluate the effectiveness of the dynamic updating semantic cache and perform sensitivity analysis on the updating threshold~$\vartheta$. As shown in Tab.~\ref{tab:hyper2}, the online semantic cache facilitates more comprehensive extraction of open-world semantics from the scene, while also enabling more effective query-based retrieval of semantic features. Our method also demonstrates strong robustness across different updating thresholds, where ambiguous semantics are implicitly corrected during the multi-view reconstruction and understanding.

\textbf{Importance of multi-level knowledge feature fusion.}
The integrated knowledge features contribute to a more comprehensive understanding of multi-level semantics in the scene by fusing representations from Grounded-SAM and CLIP. We validate the effectiveness of this module by ablating each feature extractor individually. Results reveal that relying on a single visual foundation model often introduces ambiguity: CLIP features overemphasize high-frequency regions, hindering accurate instance localization, while DINO and SAM focus on low-frequency structures, often missing fine-grained object details. Ablation results indicate that multi-level feature fusion contributes to more comprehensive and finer semantic feature extraction.

\begin{figure}[t] 
\centering
\includegraphics[width=1.0\textwidth]{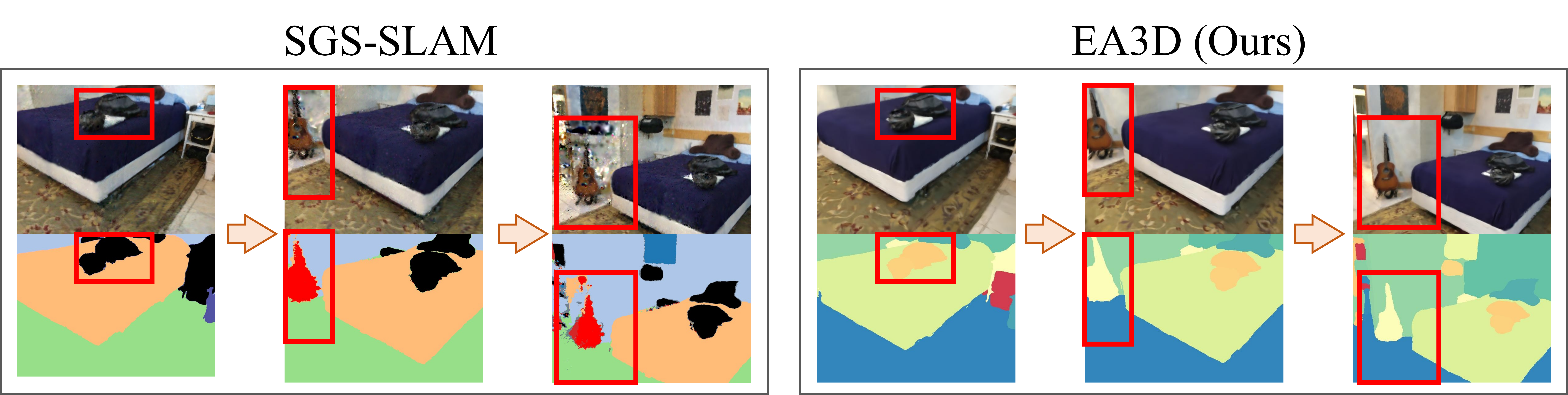}
\caption{\textbf{Compare \ourmethod{} with traditional SLAM-based methods.} 
}
\label{slam}
\end{figure}

\textbf{Importance of dense online visual odometry.}
Online visual odometry facilitates the generation of high-quality initial poses and relatively dense point cloud priors.
We validate the effectiveness of the online visual odometry by replacing the dense odometry point cloud used in our method with a sparse point cloud estimated from SfM (Colmap), a commonly used offline pose estimator in traditional 3D reconstruction and understanding pipelines. Experimental results indicate that a sparse initial point cloud from SfM results in unreliable keypoint matching and struggles to capture fine-scale structures or small objects. In contrast, our method leverages a fused dense point cloud from Cut3R~\cite{wang2025continuous}, enabling more accurate and timely reconstruction of detailed geometry.

\textbf{Design of online feature Gaussian.} To enable online streaming scene reconstruction and understanding, we propose a strategy based on online Gaussian feature optimization. Existing online reconstruction methods can be roughly categorized into two types: 1) StreamGS-based approaches~\cite{sun20243dgstream, gao2024hicom, li2025streamgs}, which require multi-view video streams and pre-defined camera poses, and 2) Online SLAM-based methods~\cite{li2025monogs++, matsuki2024gaussian}, which struggle with modeling dynamic movements and fine-grained geometry, and rely heavily on expensive post-refinement for satisfactory reconstruction and rendering quality. More critically, both types of methods lack scene understanding capabilities and are unable to capture object-level semantics and geometry in an online manner. Inspired by both paradigms, we propose an online framework that enables real-time reconstruction of large-scale environments and fine-grained object geometry, while simultaneously inferring semantic information. \ourmethod{} integrates SLAM-based online pose estimation and further enhances the reconstruction of complex objects via dense, per-pixel geometry modeling and feature embedding. In contrast to HiCoM~\cite{gao2024hicom}, our baseline method, \ourmethod{} operates without relying on external pose priors or multi-view streaming input, enabling plug-and-play scene reconstruction in dynamic environments and offering greater applicability in real-world scenarios.

\textbf{Effectiveness of regularization term.} We further ablate the effect of semantic-awareness regularization term $\mathcal{L}_{\mathbf{s}}$ by removing it. Ablation shows that regularization term facilitates both geometric reconstruction and rendering quality, which helps optimize instance-level Gaussian distributions, thereby better promoting the joint optimization of semantic knowledge and scene geometry.

\textbf{Robustness under challenging conditions.} To further quantify the robustness and accuracy of our model under various challenging conditions, we thoroughly collected scenes and video clips from the benchmark that feature severe occlusion, rapid camera motion, and simulated low-texture environments. Targeted validation experiments were conducted on these challenging cases, as presented in the Table~\ref{tab:chanllenge}. Our method demonstrates outstanding robustness and accuracy, surpassing the baseline approaches even under such difficult conditions.

\setlength\tabcolsep{8.0pt}
\begin{table*}
    \small
    \centering
    \captionsetup{type=table}
    \caption{Robustness evaluation under challenging conditions, including severe occlusion, rapid camera motion, and low-texture environments. ``Rec.'' reflects the accuracy of online geometry and visual odometry, while ``Seg.'' represents multi-view 3D understanding, which can be used to assess semantic coherence.}
    \begin{tabular}{ lccccc }
        \toprule
        Methods & Occlusion & Fast motion & Low-texture\\
        \midrule
        Baseline (Rec.) & 18.4 & 20.2 & 22.3  \\
        \textbf{Ours (Rec.)} & 23.1 & 23.3 & 22.9  \\
        Baseline (Seg.) & 31.6 & 34.8 & 35.8  \\
        \textbf{Ours (Seg.)} & 39.5 & 41.1 & 44.3  \\
        \bottomrule
    \end{tabular}
    \label{tab:chanllenge}
    \vspace{-10pt}
\end{table*}

\textbf{Hyperparameters.}
We provide a further ablation study on the hyperparameters in our method, including the loss weight balancing factors $\lambda_1$, $\lambda_2$, and $\lambda_3$, odometry update threshold $\varrho$, pruning threshold $\zeta$, and the semantic cache updating threshold $\vartheta$. As shown in Tab.~\ref{tab:hyper1} and Tab.~\ref{tab:hyper2}, our method demonstrates strong robustness to hyperparameter variations.

\setlength\tabcolsep{8.0pt}
\begin{table*}
    \small
    \centering
    \captionsetup{type=table}
    \caption{Ablation on hyperparameters $\lambda_1$, $\lambda_2$, and $\lambda_3$.}
    \begin{tabular}{ lccccc }
        \toprule
        $\lambda_1$ & $\lambda_2$ & $\lambda_3$ & Rec.(PSNR $\uparrow$) & Seg.(mIoU $\uparrow$)\\
        \midrule
        0.10 & 0.25 & 0.10 & 25.7 & 45.9 \\
        0.15 & 0.15 & 0.20 & 25.3 & 46.3 \\
        0.20 & 0.20 & 0.20 & 25.6 & 46.0 \\
        0.25 & 0.10 & 0.15 & 25.8 & 46.3 \\
        \bottomrule
    \end{tabular}
    \label{tab:hyper1}
    \vspace{-5pt}
\end{table*}

\setlength\tabcolsep{8.0pt}
\begin{table*}
    \small
    \centering
    \captionsetup{type=table}
    \caption{Ablation on hyperparameters, including odometry update threshold $\varrho$, pruning threshold $\zeta$, and the semantic cache updating threshold $\vartheta$.}
    \begin{tabular}{ lcc|ccc|ccc }
        \toprule
        $\varrho$ & PSNR $\uparrow$ & mIoU $\uparrow$ & $\zeta$ & PSNR $\uparrow$ & mIoU $\uparrow$ & $\vartheta$ & PSNR $\uparrow$ & mIoU $\uparrow$ \\
        \midrule
        0.5 & 24.9 & 45.8 & $2 \times 10^{-2}$ & 24.8 & 45.6  & 0.5 & 25.4 & 45.9  \\
        0.6 & 25.4 & 46.1 & $2 \times 10^{-3}$ & 25.5 & 45.9  & 0.6 & 25.8 & 46.3  \\
        0.7 & 25.8 & 46.3 & $2 \times 10^{-4}$ & 25.8 & 46.3  & 0.7 & 25.7 & 45.4  \\
        0.8 & 25.6 & 46.0 & $2 \times 10^{-5}$ & 25.0 & 46.1  & 0.8 & 25.8 & 45.0  \\
        \bottomrule
    \end{tabular}
    \label{tab:hyper2}
\end{table*}

\begin{figure}[t] 
\centering
\includegraphics[width=1.0\textwidth]{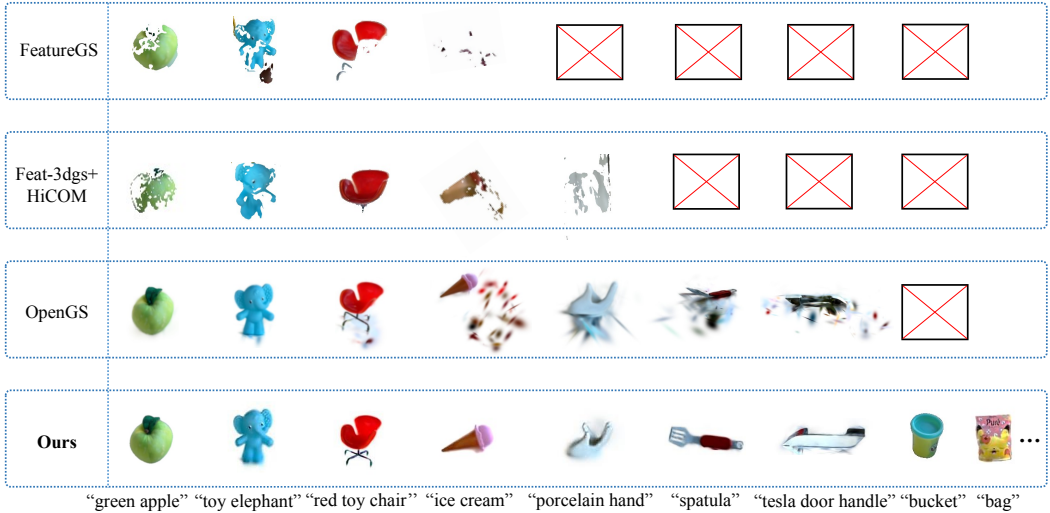}
\caption{\textbf{Visualization of 3D Object Extraction.} 
}
\label{fig:lerf-compare}
\vspace{-15pt}
\end{figure}

\section{More Qualitative Visualizations}
\label{visual}
We provide additional qualitative visual comparisons as shown in Fig.~\ref{fig:lerf-compare} and Fig.~\ref{slam}. Results demonstrate that our online feature Gaussians capture both geometric structure and semantic context with remarkable efficiency and precision. By jointly optimizing for pose estimation and feature representation, our method produces coherent, high-fidelity reconstructions that preserve fine details and semantic consistency. In contrast, state-of-the-art baselines often suffer from noisy feature aggregation, leading to degraded rendering quality and a failure to recognize or reconstruct complex or ambiguous objects—particularly those with limited observations or underrepresented categories.

\section{Diverse Downstream Applications}
\label{application}

\begin{figure}[t] 
\centering
\includegraphics[width=1.0\textwidth]{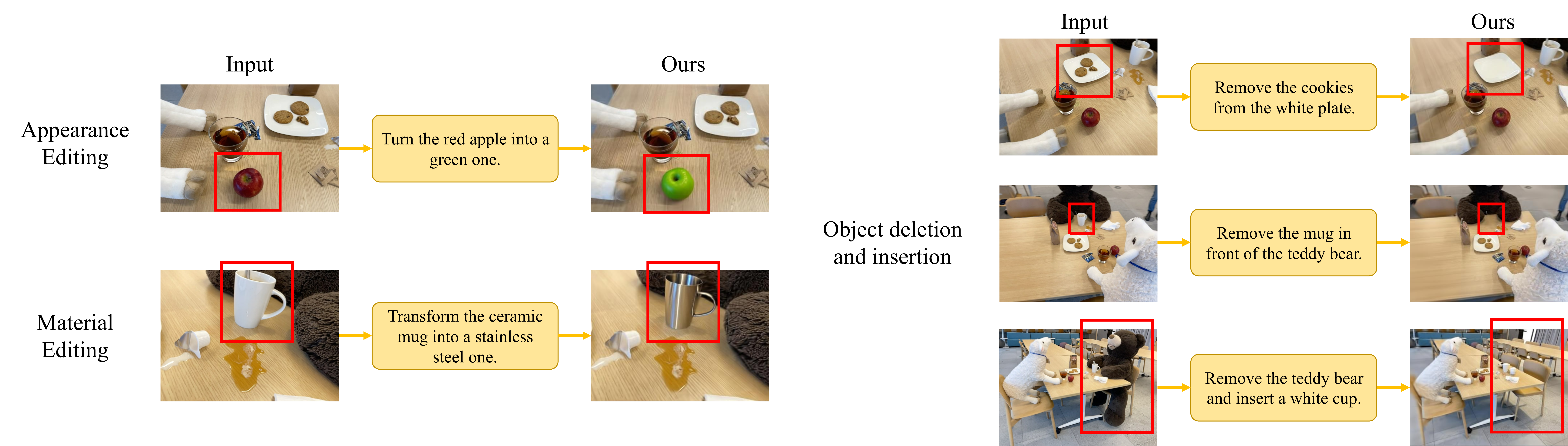}
\caption{\textbf{Visualization of Diverse Downstream Applications.} 
}
\label{fig:lerf-edit}
\vspace{-15pt}
\end{figure}

\ourmethod{} facilitates diverse downstream applications by dynamically aligning with LLM instructions or text-to-image generation models. As illustrated in Fig.~\ref{fig:lerf-edit}, combining \ourmethod{} with controllable generation and editing enables compelling functionalities such as manipulation simulation, motion emulation, controllable 3D editing, and object insertion or removal.

\section{Failure Cases and Limitations}
\label{failure}
The primary limitation of our approach arises from the imperfect accuracy and completeness of semantic extraction by vision-language models (VLMs) and vision foundation models (VFMs) in open-world scenarios. In particular, when VLMs generate incorrect semantic interpretations, our method may struggle to fully rectify these errors, leading to semantic mismatches within certain geometric regions. Although our approach supports implicit semantic alignment and correction, it fails to reconstruct geometry or resolve semantic ambiguity for objects that appear in only a few frames (e.g., a single frame in the entire video). This limitation is inherent to the underlying principles of multi-view reconstruction. In future work, we plan to integrate autoregressive and diffusion-based generative models to enable robust geometric and semantic reasoning under single-view or severely occluded conditions.

\section{Broader Impacts}
\label{impact2}
This paper presents research aimed at advancing the fields of 3D vision, which hold significant promise for enhancing the 3D object extraction. While AI-driven scene reconstruction and perception bring benefits, they could also raise concerns regarding their social and economic impacts. Automating 3D labeling and perception tasks can potentially disrupt the labor market, posing risks to certain job sectors, particularly in sectors that rely on manual data annotation. It is crucial to exercise caution and ensure that the societal implications are thoroughly addressed.


\end{document}